\documentclass[final]{cvpr}

\usepackage{times}
\usepackage{epsfig}
\usepackage{graphicx}
\usepackage{amsmath}
\usepackage{amssymb}

\usepackage{subfigure}
\usepackage{multicol,multicap}
\usepackage{algorithm}
\usepackage{algorithmic}
\usepackage{booktabs,tabularx,multirow}

\usepackage[pagebackref=true,breaklinks=true,colorlinks,bookmarks=false]{hyperref}

\def\cvprPaperID{368} % *** Enter the CVPR Paper ID here
\def\confYear{CVPR 2021}

\begin{document}

%%%%%%%%% TITLE
\title{R$^3$Det\bf{$:$} Refined Single-Stage Detector with Feature Refinement \\for Rotating Object}

\author{Xue Yang$^{1,2}$, Junchi Yan$^{1,2}$\thanks{Corresponding author is Junchi Yan.}, Ziming Feng$^{3}$, Tao He$^{4}$\\
$^{1}$Department of Computer Science and Engineering, Shanghai Jiao Tong University\\
$^{2}$MoE Key Lab of Artificial Intelligence, AI Institute, Shanghai Jiao Tong University\\
$^{3}$China Merchants Bank Credit Card Center\\
$^{4}$Anhui COWAROBOT CO., Ltd.\\
{\tt\small \{yangxue-2019-sjtu, yanjunchi\}@sjtu.edu.cn} \\
{\tt\small zimingfzm@cmbchina.com \quad Tommie.he@cowarobot.com}
}

\maketitle

%%%%%%%%% ABSTRACT
\begin{abstract}
   Rotation detection is a challenging task due to the difficulties of locating the multi-angle objects and separating them effectively from the background. Though considerable progress has been made, for practical settings, there still exist challenges for rotating objects with large aspect ratio, dense distribution and category extremely imbalance. In this paper, we propose an end-to-end refined single-stage rotation detector for fast and accurate object detection by using a progressive regression approach from coarse to fine granularity. Considering the shortcoming of feature misalignment in existing refined single-stage detector, we design a feature refinement module to improve detection performance by getting more accurate features. The key idea of feature refinement module is to re-encode the position information of the current refined bounding box to the corresponding feature points through pixel-wise feature interpolation to realize feature reconstruction and alignment. For more accurate rotation estimation, an approximate SkewIoU loss is proposed to solve the problem that the calculation of SkewIoU is not derivable. Experiments on three popular remote sensing public datasets DOTA, HRSC2016, UCAS-AOD as well as one scene text dataset ICDAR2015 show the effectiveness of our approach. Tensorflow and Pytorch version codes are available at \url{https://github.com/Thinklab-SJTU/R3Det_Tensorflow} and \url{https://github.com/SJTU-Thinklab-Det/r3det-on-mmdetection}, and R$^3$Det is also integrated in our open source rotation detection benchmark: \url{https://github.com/yangxue0827/RotationDetection}.
\end{abstract}

%%%%%%%%% BODY TEXT
\section{Introduction}
	Object detection is one of the fundamental tasks in computer vision, and many high-performance general-purpose object detectors have been proposed. Current popular detection methods can be in general divided into two types: two-stage object detectors \cite{girshick2014rich,girshick2015fast,ren2015faster,dai2016r,lin2017feature} and single-stage object detectors \cite{liu2016ssd,redmon2016you,lin2017focal}. Two-stage methods have achieved promising results on various benchmarks, while the single-stage approach maintains faster detection speed.
	
\begin{figure}[!tb]
		\centering
		\subfigure[RetinaNet-Reg]{
			\begin{minipage}[t]{0.46\linewidth}
				\centering
				\includegraphics[width=0.98\linewidth]{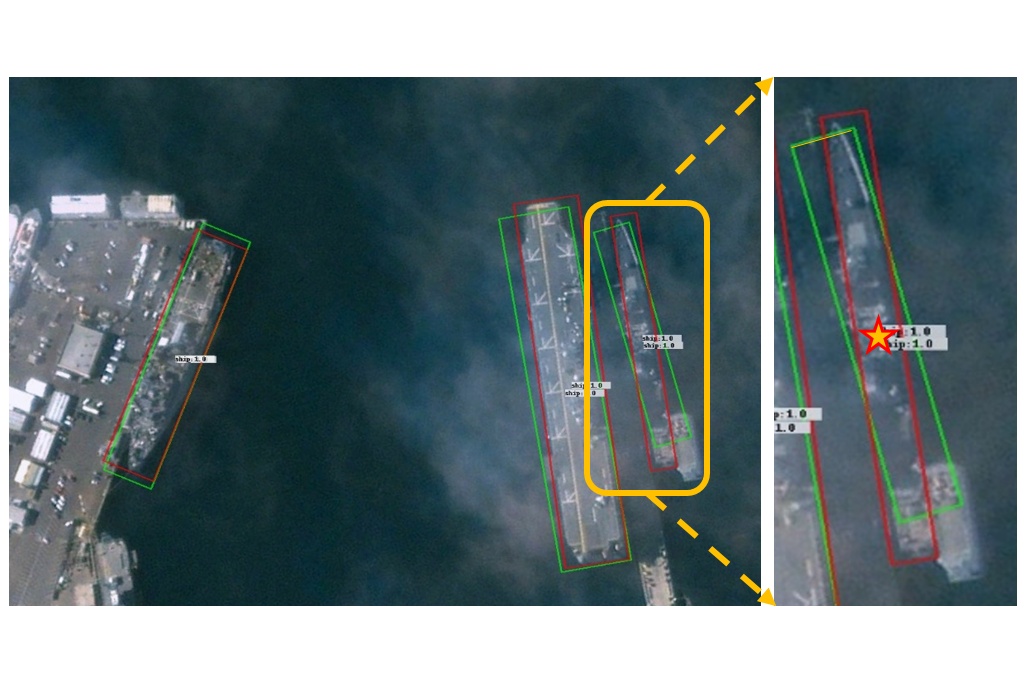}
			\end{minipage}%
			\label{fig:ship_iou}
		}
		\subfigure[RetinaNet-CSL]{
			\begin{minipage}[t]{0.46\linewidth}
				\centering
				\includegraphics[width=0.98\linewidth]{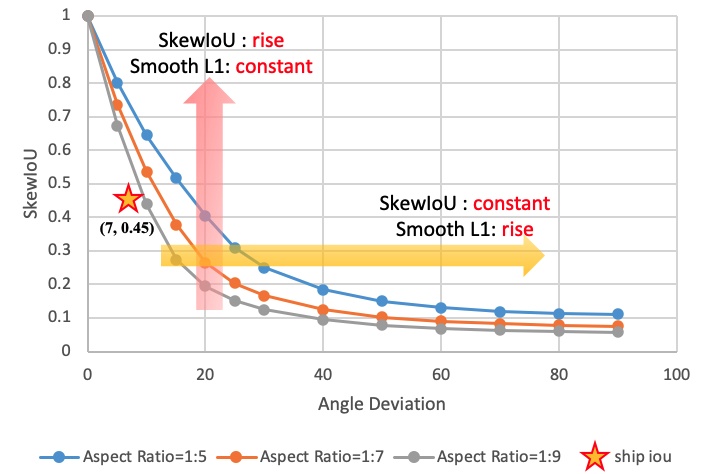}
			\end{minipage}%
			\label{fig:iou_angle}
		}
		\vspace{-5pt}
		\caption{The SkewIoU scores vary with the angle deviation. The red and green rectangles represent the ground truth and the prediction bounding box, respectively.}
		\label{fig:statistics}
		\vspace{-10pt}
	\end{figure}

	However, current general horizontal detectors have fundamental limitations for many practical applications. For instance, scene text detection, retail scene detection and remote sensing object detection whereby the objects can appear in various orientations. Therefore, many rotation detectors based on a general detection framework have been proposed in the above fields. In particular, three challenges are pronounced, as analyzed as follows:
	
	1) \textbf{Large aspect ratio.} The Skew Intersection over Union (SkewIoU) score between large aspect ratio objects is sensitive to change in angle, as sketched in Figure \ref{fig:iou_angle}.
	
	2) \textbf{Densely arranged.} As illustrated in Figure \ref{fig:anchor_vis}, many objects usually appear in densely arranged forms.
	
	3) \textbf{Arbitrary orientations.} Objects in images can appear in various orientations, which requires the detector to have accurate direction estimation capabilities.
	
	This paper is devoted to design an accurate and fast rotation detector. To maintain high detection accuracy and speed for large aspect ratio objects, we have adopted a refined single-stage rotation detector. \textbf{First}, we find that rotating anchors can perform better in dense scenes, while horizontal anchors can achieve higher recalls in fewer quantities. Therefore, a progressive regression form from coarse to fine is adopted in the refined single-stage detector, that is, the horizontal anchors are used in the first stage for faster speed and higher recall, and then the refined rotating anchors are used in the subsequent refinement stages to adapt to intensive scenarios. \textbf{Second}, we also notice that existing refined single-stage detectors \cite{Zhang2018Single,chi2019selective} have feature misalignment problems\footnote{Mainly refers to misalignment between region of interest (RoI) and the feature, see details in Figure \ref{fig:refine_3}.}, which greatly limits the reliability of classification and regression during the refined stages. We design a feature refinement module FRM that uses the feature interpolation to obtain the position information correspond to the refined anchors and reconstruct the whole feature map by pixel-wise manner to achieve feature alignment. FRM can also reduce the number of refined bounding box after the first stage, thus speeding up the model. Experimental results have shown that feature refinement is sensitive to location and its improvement in detection results is very noticeable. \textbf{Finally}, an approximate SkewIoU loss is devised to address the indifferentiable problem of SkewIoU calculation for more accurate rotation estimation. Combing these three techniques as a whole, our approach achieves state-of-the-art performance with considerable speed on four public rotating sensitive datasets including DOTA, HRSC2016, UCAS-AOD, and ICDAR2015. Specifically, this work makes the following contributions:
	
	\begin{figure}[!tb]
		\begin{center}
			\includegraphics[width=0.95\linewidth]{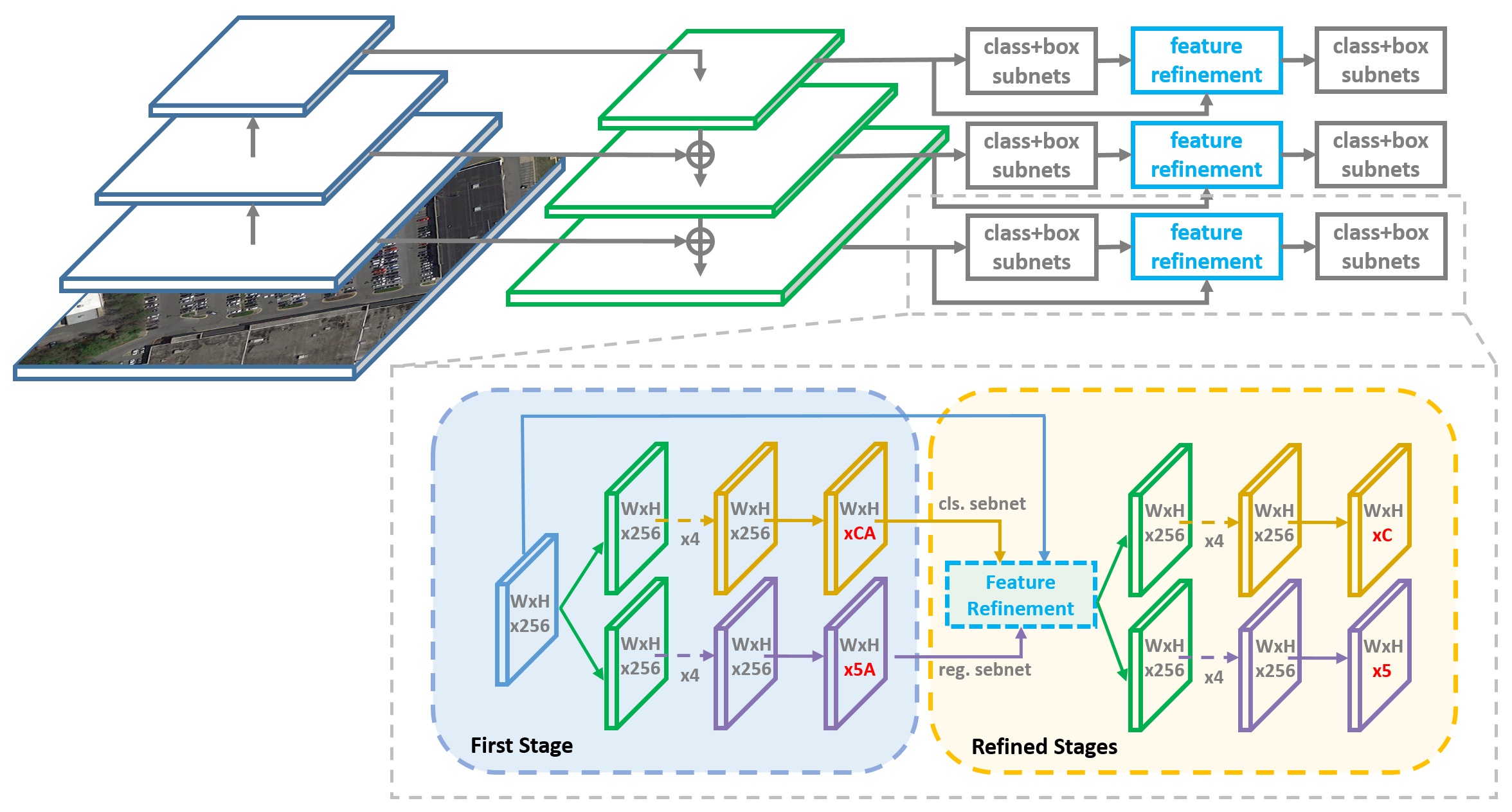}
		\end{center}
		\vspace{-8pt}
		\caption{The architecture of the proposed Refined Rotation Single-Stage Detector (RetinaNet as an embodiment). The refinement stage can be repeated by multiple times. 'A' indicates the number of anchors on each feature point, and 'C' indicates the number of categories.}
		\label{fig:pipeline}
		\vspace{-10pt}
	\end{figure}
	
	1) For large aspect ratio object detection, an accurate and fast rotation singe-stage detector is devised in a refined manner, for high-precision detection. In contrast to the recent learning based methods \cite{chen2018dual,jang2019propose,zhang2019cascade} for feature alignment, which lacks an explicit mechanism to compensate the misalignment, we propose a direct and effective pure computing based approach which is further extended to handle the rotation case. To our best knowledge, it is the first work for solving the feature misalignment problem for rotation detection.
	
	2) For densely arranged objects, we develop an efficient coarse-to-fine progressive regression approach to better exploring the two forms of anchors in a more flexible manner, tailored to each detection stage. Compared with the previous methods \cite{ma2018arbitrary,yang2018automatic,fu2018ship,yang2018object,yang2019building} using one single anchor form, our method is more flexible and efficient.
	
	3) For arbitrarily-rotated objects, a derivable approximate SkewIoU loss is devised for more accurate rotation estimation.
	Compared with the over-approximation of SkewIoU loss in recent work \cite{chen2020piou}, our method retains the accurate SkewIoU amplitude and only approximates the gradient direction of SkewIoU loss.

\section{Related Work}
	\noindent \textbf{Two-Stage Object Detectors.}
	Most existing two-stage methods are region based. In a region based framework, category-independent region proposals are generated from an image in the first stage, followed with feature extraction from these regions, and then category-specific classifiers and regressors are used for classification and regression in the second stage. Finally, the detection results are obtained by using post-processing methods such as non-maximum suppression (NMS). Faster-RCNN \cite{ren2015faster}, R-FCN \cite{dai2016r}, and FPN \cite{lin2017feature} are classic structures in a two-stage approach that can detect object quickly and accurately in an end-to-end manner. 
	
	\begin{figure}[!tb]
		\begin{center}
			\includegraphics[width=0.8\linewidth]{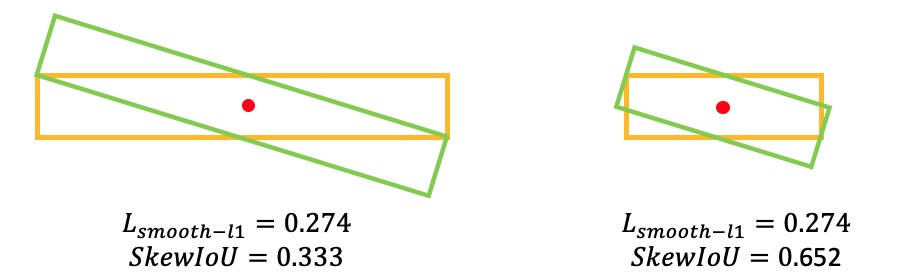}
		\end{center}
		\vspace{-8pt}
		\caption{Comparison between SkewIoU and Smooth L1 Loss.}
		\label{fig:iou_smooth-l1}
		\vspace{-10pt}
	\end{figure}
	
	\noindent \textbf{Single-Stage Object Detectors.}
	For their efficiency, single-stage detection methods are receiving more and more attention. OverFeat \cite{sermanet2013overfeat} is one of the first single-stage detectors based on convolutional neural networks. It performs object detection in a multiscale sliding
	window fashion via a single forward pass through the CNN. Compared with region based methods, Redmon et al. \cite{redmon2016you} propose YOLO, a unified detector casting object detection as a regression problem from image pixels to spatially separated bounding boxes and associated class probabilities. To preserve real-time speed without sacrificing too much detection accuracy, Liu et al. \cite{liu2016ssd} propose SSD. The work \cite{lin2017focal} solves the class imbalance problem by proposing RetinaNet with Focal loss and further improves the accuracy of single-stage detector.
	
	\noindent \textbf{Rotation Object Detectiors.}
	Remote sensing, scene text and retail scene are the main application scenarios of the rotation detector. Due to the complexity of the remote sensing image scene and the large number of small, cluttered and rotated objects, two-stage rotation detectors are still dominant for their robustness. Among them, ICN \cite{azimi2018towards}, ROI-Transformer \cite{ding2018learning}, SCRDet \cite{yang2019scrdet} and Gliding Vertex \cite{xu2020gliding} are state-of-the-art detectors. However, they use a more complicated structure causing speed bottleneck. For scene text detection, there are many efficient rotation detection methods, including both two-stage methods (R$^2$CNN \cite{jiang2017r2cnn}, RRPN \cite{ma2018arbitrary}, FOTS \cite{liu2018fots}), as well as single-stage methods (EAST \cite{Zhou2017EAST}, TextBoxes++ \cite{liao2018textboxes++}). For retail scene detection, DRN \cite{pan2020dynamic}  and PIoU \cite{chen2020piou} Loss are the latest two rotation detectors used in retail scene detection, and two rotation retail datasets are proposed respectively.
	
	\noindent \textbf{Refined Object Detectors.}
	To achieve better detection accuracy, many cascaded or refined detectors are proposed. The Cascade RCNN \cite{cai2018cascade}, HTC \cite{chen2019hybrid}, and FSCascade \cite{li2019rethinking} perform multiple classifications and regressions in the second stage, which greatly improved the detection accuracy. The same idea is also used in single-stage detectors, such as RefineDet \cite{Zhang2018Single}. Unlike the two-stage detectors, which use RoI Pooling \cite{girshick2015fast} or RoI Align \cite{he2017mask} for feature alignment. The currently refined single-stage detector is not well resolved in this respect. An important requirement of the refined single-stage detector is to maintain a full convolutional structure, which can retain the advantage of speed, but methods such as RoI Align cannot satisfy it whereby fully-connected layers have to be introduced. Although some works \cite{chen2018dual,jang2019propose,zhang2019cascade} use deformable convolution \cite{dai2017deformable} for feature alignment, whose offset parameters are often obtained by learning the offset between the pre-defined anchor box and the refined anchor. These deformable-based feature alignment methods are too implicit and can not ensure that features are truely aligned. Feature misalignment still limits the performance of the refined single-stage detector. Compared to these methods, our method can clearly find the corresponding feature area by calculation and achieve the purpose of feature alignment by feature map reconstruction.
	
\section{The Proposed Method}
	We give an overview of our method as sketched in Figure \ref{fig:pipeline}. The embodiment is a refined single-stage rotation detector based on the RetinaNet \cite{lin2017focal}, namely Refined Rotation RetinaNet (R$^3$Det). The refinement stage (which can be added and repeated by multiple times) is added to the network to refine the bounding box, and the feature refinement module FRM is added during the refinement stage to reconstruct the feature map. In a single-stage rotating object detection task, continuous refinement of the predicted bounding box can improve the regression accuracy, and feature refinement is a necessary process for this purpose.
	
	\subsection{Rotation RetinaNet}
	\noindent \textbf{Base Setting.} 
	RetinaNet is one of the most advanced single-stage detectors available today. It consists of two parts: backbone network, classification and regression subnetwork. 
	%RetinaNet adopts the Feature Pyramid Network (FPN) \cite{lin2017feature} as the backbone network. Each layer of the FPN is connected to a classification subnet and a regression subnet for predicting categories and locations. Note that the object classification subnet and the box regression subnet, though sharing a common structure, use separate parameters. RetinaNet has proposed focal loss~ \cite{lin2017focal} to address the problem caused by category imbalance, which has greatly improved the accuracy of single-stage detector.
	For RetinaNet-based rotation detection, we use five parameters ($x,y,w,h,\theta$) to represent arbitrary-oriented rectangle. Ranging in $[-\pi/2,0)$, $\theta $ denotes the acute angle to the x-axis, and for the other side we refer it as $w$. Therefore, it calls for predicting an additional angular offset in the regression subnet, whose rotation bounding box is:
	\begin{equation}
	\begin{aligned}
	t_{x}&=(x-x_{a})/w_{a}, t_{y}=(y-y_{a})/h_{a} \\
	t_{w}&=\log(w/w_{a}), t_{h}=\log(h/h_{a}), t_{\theta}=\theta-\theta_{a}\\
	t_{x}^{'}&=(x_{}^{'}-x_{a})/w_{a}, t_{y}^{'}=(y_{}^{'}-y_{a})/h_{a} \\
	t_{w}^{'}&=\log(w_{}^{'}/w_{a}), t_{h}^{'}=\log(h_{}^{'}/h_{a}), t_{\theta}^{'}=\theta_{}^{'}-\theta_{a}
	\label{eq:regression2}
	\end{aligned}
	\end{equation}
	where $x,y,w,h,\theta$ denote the box's center coordinates, width, height and angle, respectively. Variables $x, x_{a}, x^{'}$ are for the ground-truth box, anchor box, and predicted box, respectively (likewise for $y,w,h,\theta$).
	
	\noindent \textbf{Loss Function.} 
	As shown in Figure \ref{fig:iou_smooth-l1}, each box set has the same center point, height and width. The angle difference between the two box sets is the same, but the aspect ratio is different. As a result, the smooth L1 loss value of the two sets is the same (mainly from the angle difference), but the SkewIoU is quite different. The red and orange arrows in Figure \ref{fig:iou_angle} show the inconsistency between SkewIoU and smooth L1 Loss. We can draw conclusion that smooth L1 loss function is still not suitable for rotation detection, especially for objects with large aspect ratios, which are sensitive to SkewIoU. What's more, the evaluation metric of rotation detection is also dominated by SkewIoU.
	
	The IoU related loss is an effective regression loss function that can solve above problem and is already widely used in horizontal detection, such as GIoU \cite{rezatofighi2019generalized}, DIoU \cite{zheng2020distance}, etc. However, the SkewIoU calculation function between two rotating boxes is underivable, which means that we cannot directly use the SkewIoU as the regression loss function. Inspired by SCRDet \cite{yang2019scrdet}, we propose a derivable approximate SkewIoU loss, the multi-task loss is defined as follows:
	\begin{equation}
	\begin{aligned}
	L  =& \frac{\lambda_{1}}{N}\sum_{n=1}^{N}obj_{n}\frac{L_{reg}(v_{n}^{'},v_{n})}{|L_{reg}(v_{n}^{'},v_{n})|}|f(SkewIoU)| \\
	& + \frac{\lambda_{2}}{N}\sum_{n=1}^{N}L_{cls}(p_{n},t_{n})
	\label{eq:multitask_loss}
	\end{aligned}
	\end{equation}
	\begin{equation}
	\begin{aligned}
	L_{reg}(v^{'},v) = L_{smooth-l1}(v_{\theta}^{'},v_{\theta}) - IoU(v_{\{x,y,w,h\}}^{'}, v_{\{x,y,w,h\}})
	\label{eq:reg_iou_loss}
	\end{aligned}
	\end{equation}
% 	\begin{equation}
% 	\begin{aligned}
% 	L & = \frac{\lambda_{1}}{N}\sum_{n=1}^{N}obj_{n}\sum_{j\in\{x,y,w,h,\theta\}}\frac{L_{reg}(v_{nj}^{'},v_{nj})}{|L_{reg}(v_{nj}^{'},v_{nj})|}|f(SkewIoU)| \\
% 	& + \frac{\lambda_{2}}{N}\sum_{n=1}^{N}L_{cls}(p_{n},t_{n})
% 	\label{eq:multitask_loss}
% 	\end{aligned}
% 	\end{equation}
	where $N$ indicates the number of anchors, $obj_{n}$ is a binary value ($obj_{n}=1$ for foreground and $obj_{n}=0$ for background, no regression for background). $v^{'}$ represents the predicted offset vectors, $v$ denotes the targets vector of ground-truth. While $t_{n}$ indicates the label of object, $p_{n}$ is the probability distribution of various classes calculated by sigmoid function. SkewIoU denotes the overlap of the prediction box and ground-truth. The hyper-parameter $\lambda_{1}$, $\lambda_{2}$ control the trade-off and are set to 1 by default. The classification loss $L_{cls}$ is implemented by focal loss \cite{lin2017focal}. $|.|$ is used to obtain the modulus of the vector and is not involved in gradient back propagation. $f(.)$ represents the loss function related to SkewIoU. $IoU(.)$ represents the horizontal bounding box IoU calculation function.
	
	\begin{figure}[!tb]
		\centering
		
		\subfigure[Original image]{
			\begin{minipage}[t]{0.46\linewidth}
				\centering
				\includegraphics[width=0.98\linewidth]{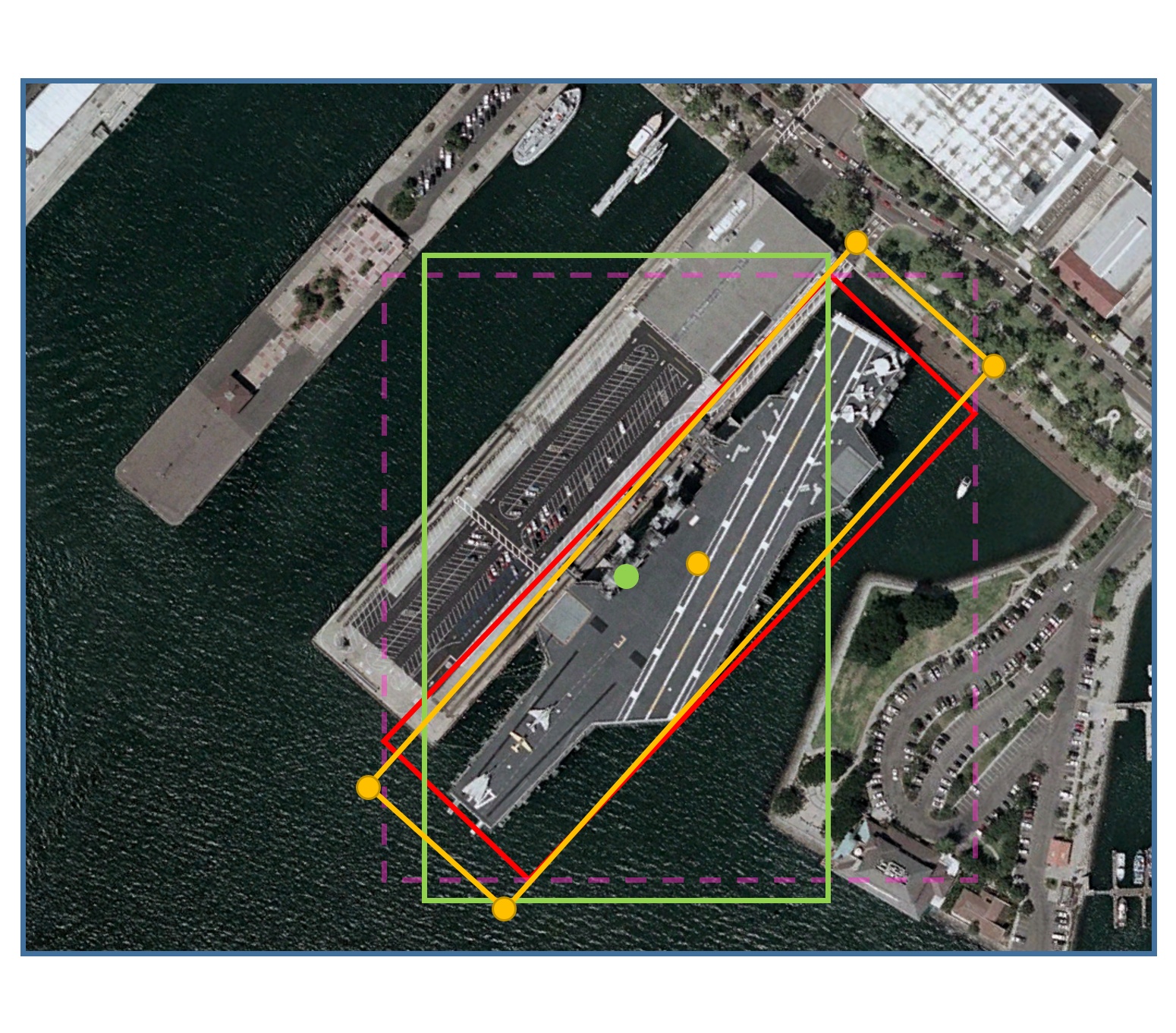}
			\end{minipage}%
			\label{fig:refine_1}
		}
		\subfigure[Feature interpolation]{
			\begin{minipage}[t]{0.46\linewidth}
				\centering
				\includegraphics[width=0.98\linewidth]{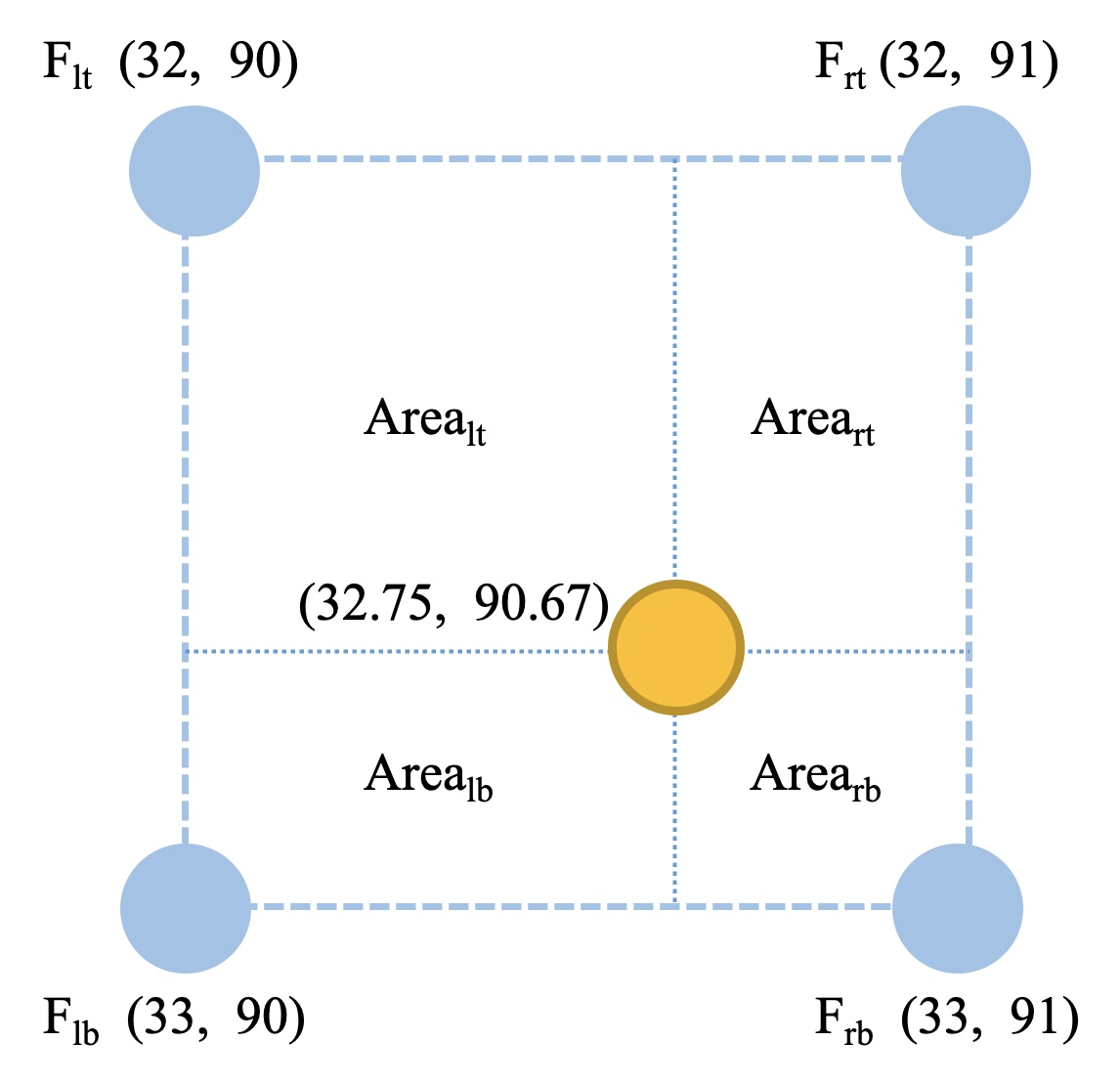}
			\end{minipage}%
			\label{fig:feature_interpolation}
		}\\
		\subfigure[Refine box with misaligned feature  due to bounding box location changes.]{
			\begin{minipage}[t]{0.46\linewidth}
				\centering
				\includegraphics[width=0.98\linewidth]{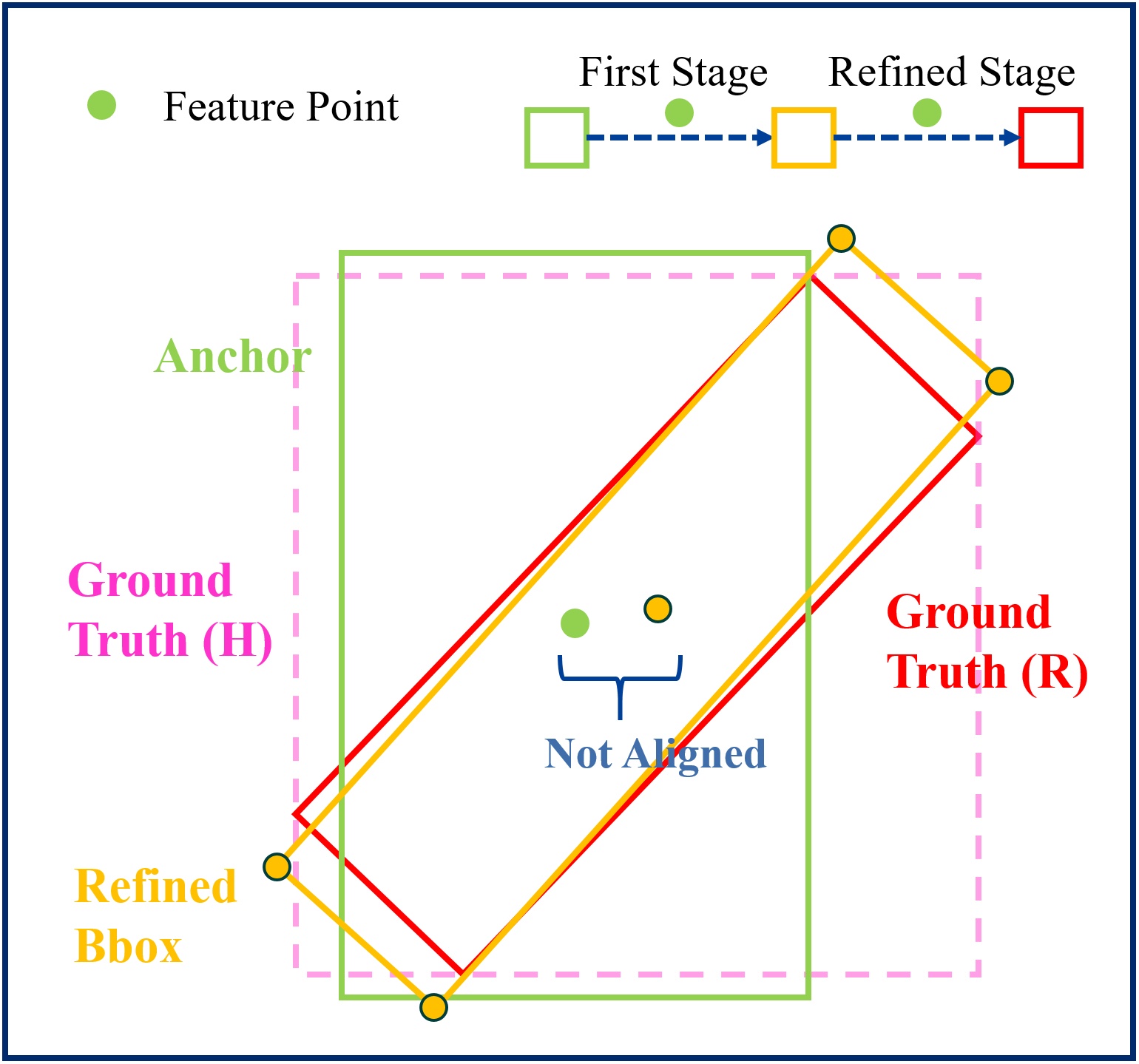}
			\end{minipage}
			\label{fig:refine_2}
		}
		\subfigure[Refine box with aligned features by reconstructing the feature map.]{
			\begin{minipage}[t]{0.46\linewidth}
				\centering
				\includegraphics[width=0.98\linewidth]{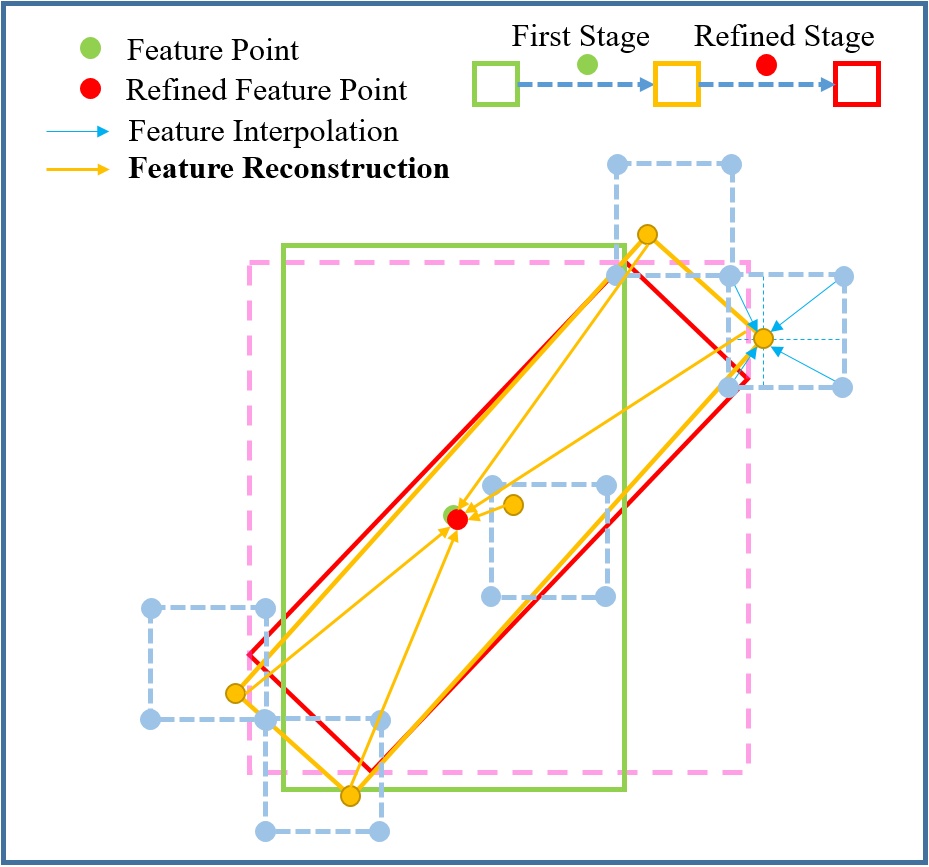}
			\end{minipage}%
			\label{fig:refine_3}
		}
        \centering
		\caption{Root cause analysis of feature misalignment and the core idea for our proposed feature refinement module.}
		\vspace{-8pt}
		\label{fig:refine}
	\end{figure}
	
	Compared to the traditional regression loss, the new regression loss can be divided into two parts, $\frac{L_{reg}(v_{n}^{'},v_{n})}{|L_{reg}(v_{n}^{'},v_{n})|}$ determines the direction of gradient propagation (a unit vector), which is an important part to ensure that the loss function is derivable. $|f(SkewIoU)|$ is responsible for adjusting the loss value (magnitude of gradient), and it is unnecessary to be derivable (a scalar). Taking into account the inconsistency between SkewIoU and smooth L1 loss, we use Equation \ref{eq:reg_iou_loss} as the dominant gradient function for regression loss. Through such a combination, the loss function is derivable, while its size is highly consistent with SkewIoU. Experiments show that the detector based on this approximate SkewIoU loss can achieve considerable gains.
	
	\begin{figure}[!tb]
		\begin{center}
			\includegraphics[width=0.85\linewidth]{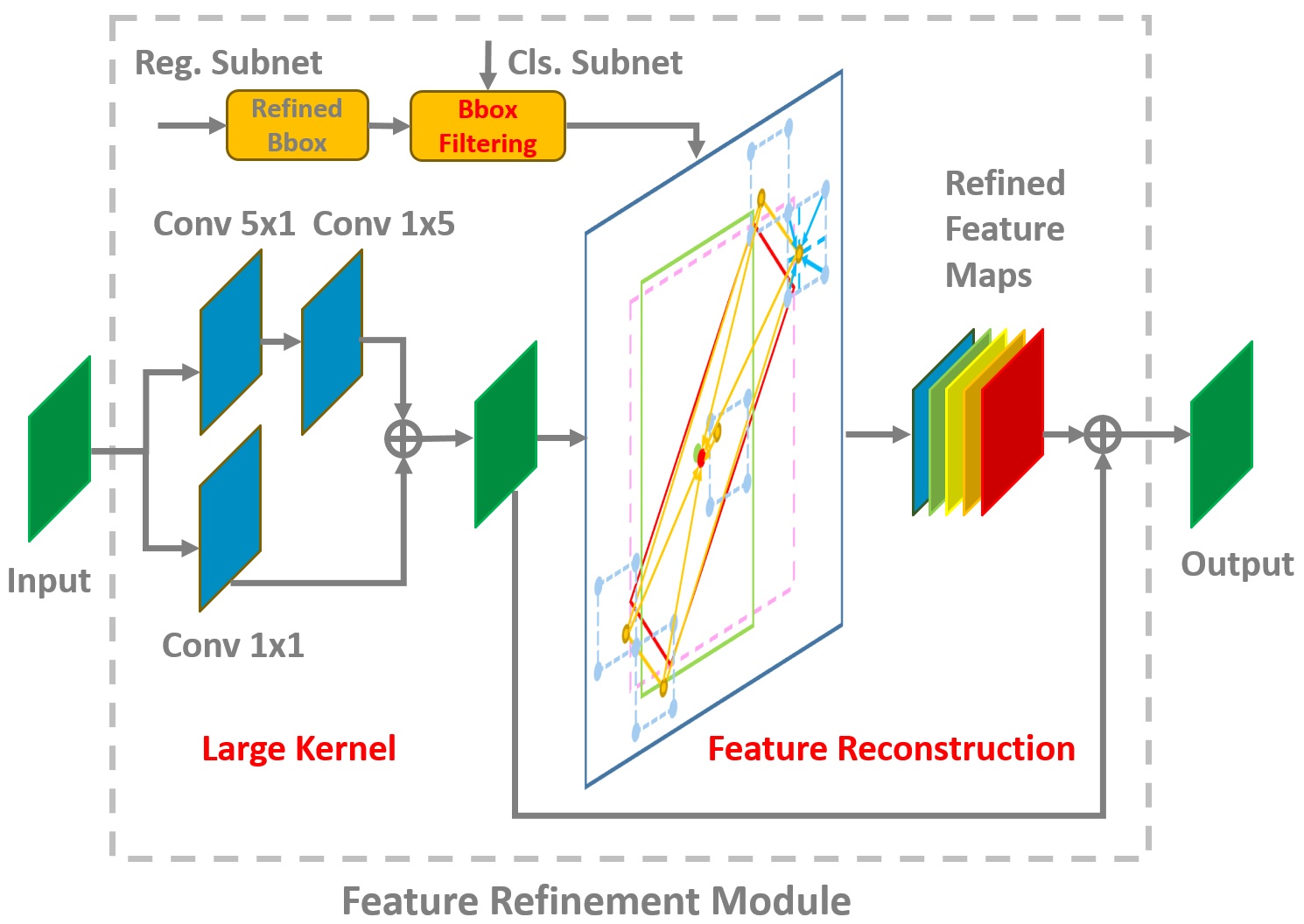}
		\end{center}
		\caption{Feature Refinement Module FRM. It mainly includes three parts: refined bounding box filtering (BF), large kernel (LK) and feature reconstruction (FR).}
		\label{fig:FRM}
		\vspace{-10pt}
	\end{figure}
		
	\subsection{Refined Rotation RetinaNet}\label{subsec:rrr}
	\textbf{Refined Detection.} 
	The SkewIoU score is sensitive to the change in angle, and a slight angle shift causes a rapid decrease in the IoU score, as shown in Figure \ref{fig:statistics}. Therefore, the refinement of the prediction box helps to improve the recall rate of the rotation detection. We join multiple refinement stages with different IoU thresholds. In addition to using the foreground IoU threshold 0.5 and background IoU threshold 0.4 in the first stage, the thresholds of first refinement stage are set 0.6 and 0.5, respectively. If there are multiple refinement stages, the remaining thresholds are 0.7 and 0.6. The overall loss for refined detector is defined as follows:
	\begin{equation}
	\begin{aligned}
	L_{total} = \sum_{i=1}^{N}\alpha_{i}L_{i}
	\label{eq:overall_loss}
	\end{aligned}
	\end{equation}
	where $L_{i}$ is the loss value of the $i$-th refinement stage and trade-off coefficients $\alpha_{i}$ are set to 1 by default.
	
% 	\begin{algorithm}[t!]
% 		\caption{Feature Refinement Module} 
% 		\label{algorithm:FRM}
% 		\hspace*{0.02in} {\bf Input:} 
% 		original feature map $F$, the bounding box ($B$) and confidence ($S$) of the previous stage\\
% 		\hspace*{0.02in} {\bf Output:} 
% 		reconstructed feature map $F'$
% 		\begin{algorithmic}[1]
% 			\State $B' \leftarrow BoxFilter(B, S)$;
% 			\State $h, w \leftarrow Shape(F)$, $F' \leftarrow ZerosLike(F)$;
% 			\State $F \leftarrow Conv_{1\times1}(F) + Conv_{1\times5}(Conv_{5\times1}(F))$
% 			\For{$i \leftarrow $ 0 \textbf{to} $ h-1 $}
% 			\For{$j \leftarrow $ 0 \textbf{to} $ w-1 $} 
% 			\State $P \leftarrow GetFivePoints(B'(i, j))$;
% 			\For{$p \in P $} 
% 			\State $p_{x} \leftarrow Min(p_{x}, w-1)$, $p_{x} \leftarrow Max(p_{x}, 0)$;
% 			\State $p_{y} \leftarrow Min(p_{y}, h-1)$, $p_{y} \leftarrow Max(p_{y}, 0)$;
% 			\State $F'(i, j) \leftarrow F'(i, j) + BilinearInte(F, p)$;
% 			\EndFor
% 			\EndFor
% 			\EndFor
% 			\State $F' \leftarrow F' + F$;
% 			\State \Return $F'$
% 		\end{algorithmic}
% 	\end{algorithm}
    \begin{figure}[!tb]
		\begin{center}
			\includegraphics[width=0.98\linewidth]{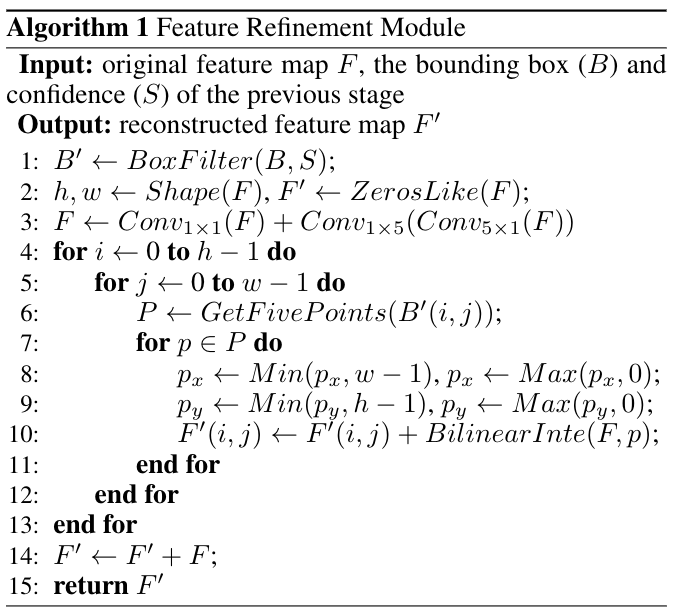}
		\end{center}
		\label{algorithm:FRM}
		\vspace{-10pt}
	\end{figure}
	
\noindent \textbf{Feature Refinement Module.} 
	Many refined detectors still use the same feature map to perform multiple classifications and regressions, without considering the feature misalignment caused by the location changes of the bounding box. Figure \ref{fig:refine_2} depicts the box refining process without feature refinement, resulting in inaccurate features, which can be disadvantageous for those categories that have a large aspect ratio or a small sample size. Here we propose to re-encode the position information of the current refined bounding box (orange rectangle) to the corresponding feature points (red point\footnote{The red and green points should be totally overlapping to each other, while here the red point is intentionally offset in order to distinguishingly visualize the entire process.}), thereby reconstructing the entire feature map by pixel-wise manner to achieve the alignment of the features. The whole process is shown in Figure \ref{fig:refine_3}. To accurately obtain the location feature information correspond to the refined bounding box, we adopt the bilinear feature interpolation method, as shown in Figure \ref{fig:feature_interpolation}. Feature interpolation can be formulated as follows:
	\begin{equation}
	\begin{aligned}
	F = &F_{lt} * A_{rb} + F_{rt} * A_{lb} + F_{rb} * A_{lt} + F_{lb} * A_{rt}
	\label{eq:overall_loss}
	\end{aligned}
	\end{equation}
	where $A$ denotes the $Area$ in Figure \ref{fig:feature_interpolation}, $F \in \mathbb{R}^{C \times 1 \times 1}$ represents the feature vector of the point on the feature map.

	Based on the above result, a feature refinement module is devised, whose structure and pseudo code is shown in Figure \ref{fig:FRM} and Algorithm \ref{algorithm:FRM}, respectively. Specifically, the feature map is added by two-way convolution to obtain a new feature (large kernel, LK). Only the bounding box with the highest score of each feature point is preserved in the refinement stage to increase the speed (box filtering, BF), meanwhile ensuring that each feature point corresponds to only one refined bounding box. The filtering of bounding boxes is a necessary step for feature reconstruction (FR). For each feature point of the feature map, we obtain the corresponding feature vector on the feature map according to the five coordinates of the refined bounding box (one center point and four corner points). A more accurate feature vector is obtained by bilinear interpolation. We add the five feature vectors and replace the current feature vector. After traversing the feature points, we reconstruct the whole feature map. Finally, the reconstructed feature map is added to the original feature map to complete the whole process.
	
	\begin{figure}[!tb]
		\centering
		\subfigure[]{
			\begin{minipage}[t]{0.30\linewidth}
				\centering
				\includegraphics[width=0.98\linewidth]{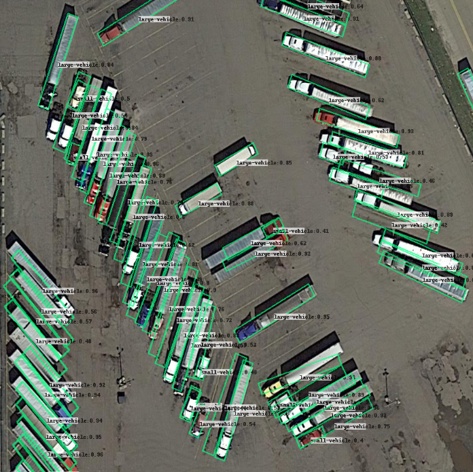}
			\end{minipage}%
			\label{fig:H_anchor_vis2}
		}
		\subfigure[]{
			\begin{minipage}[t]{0.30\linewidth}
				\centering
				\includegraphics[width=0.98\linewidth]{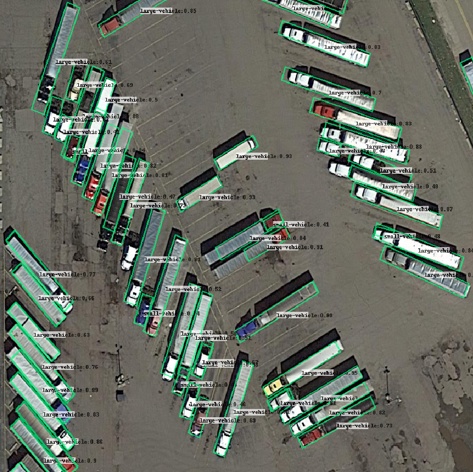}
			\end{minipage}%
			\label{fig:R_anchor_vis2}
		}
		\subfigure[]{
			\begin{minipage}[t]{0.30\linewidth}
				\centering
				\includegraphics[width=0.98\linewidth]{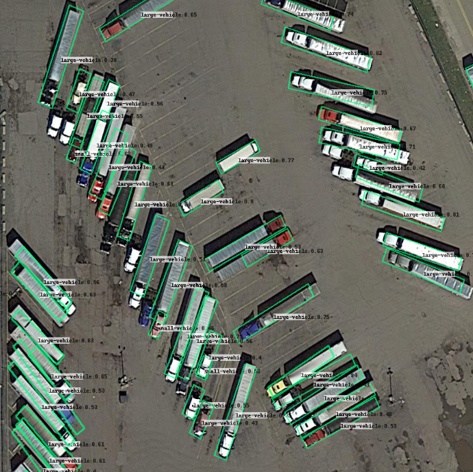}
			\end{minipage}%
			\label{fig:HR_anchor_vis2}
		}\\
		\subfigure[RetinaNet-H]{
			\begin{minipage}[t]{0.30\linewidth}
				\centering
				\includegraphics[width=0.98\linewidth]{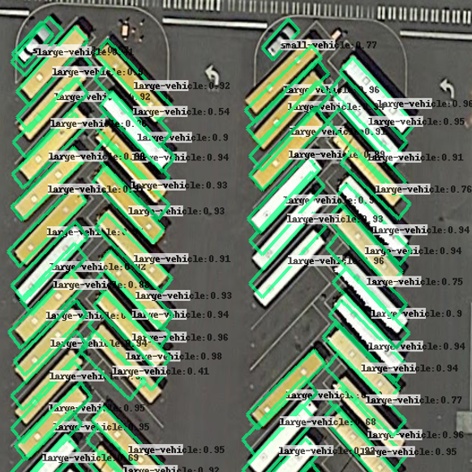}
			\end{minipage}%
			\label{fig:H_anchor_vis1}
		}
		\subfigure[RetinaNet-R]{
			\begin{minipage}[t]{0.30\linewidth}
				\centering
				\includegraphics[width=0.98\linewidth]{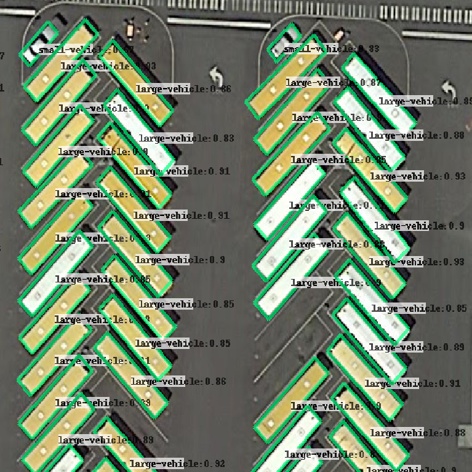}
			\end{minipage}%
			\label{fig:R_anchor_vis1}
		}
		\subfigure[R$^3$Det$^*$]{
			\begin{minipage}[t]{0.30\linewidth}
				\centering
				\includegraphics[width=0.98\linewidth]{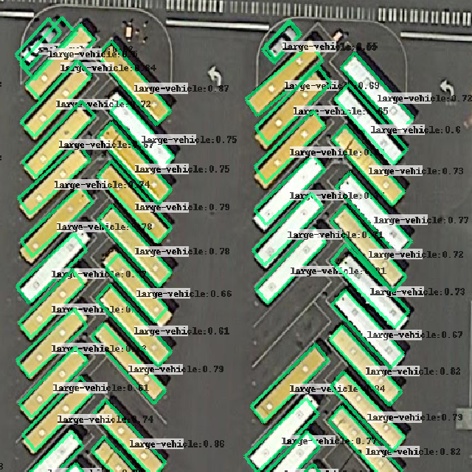}
			\end{minipage}%
			\label{fig:HR_anchor_vis1}
		}
		\caption{Visualization on DOTA. Here `H' and `R' represent the horizontal and rotating anchors, respectively. }
		\label{fig:anchor_vis}
		\vspace{-5pt}
	\end{figure}
	%-------------------------------------------------------------------------
	\begin{table}[]
		\centering
		\resizebox{0.45\textwidth}{!}{
			\begin{tabular}{lccccccc}
				
				\toprule
				\multirow{2}{*}{Method}& \multicolumn{2}{c}{FRM} &approximate&\multirow{2}{*}{SV.}&\multirow{2}{*}{LV.}&\multirow{2}{*}{SH.}&\multirow{2}{*}{mAP}\\
				
				\cmidrule(r){2-3} 
				
				& BF\&FR & LK & SkewIoU loss &\\
				\midrule
				RetinaNet-R  & & & & 64.64 & 71.01 & 68.62 & 62.76 \\
				RetinaNet-H  & & & & 63.50 & 50.68 & 65.93 & 62.79 \\
				%RetinaNet-H  & & & $\checkmark$ & 62.13 & 53.85 & 66.50 & 65.34 \\
				R$^3$Det$^*$  & & $\checkmark$ & & 65.02 & 67.31 & 67.31 & 63.52 \\
				%R$^3$Det$^*$ & ResNet101 & & $\checkmark$ & & $\checkmark$ &  \\
				\hline
				R$^3$Det   & $\checkmark$ & $\checkmark$ & & 65.81 & 72.76 & 70.14 & 66.31 \\
				R$^3$Det$^{\dag}$  & $\checkmark$& $\checkmark$ & & 67.45 & 73.98 & 70.27 & 67.66 \\
				R$^3$Det$^{\dag}$  & $\checkmark$ & $\checkmark$ &$\checkmark$ & 68.04 & 72.72 & 76.03 & 69.50\\
				%R$^3$Det$^{\dag}$ & ResNet101 & $\checkmark$ & $\checkmark$ &$\checkmark$ & & 68.34 & 74.10 & 77.14 & 70.58 \\
				%R$^3$Det$^{\dag}$ & ResNet101 & $\checkmark$ & $\checkmark$ & $\checkmark$ &  $\checkmark$ & 71.24 & 80.07 & 78.82 & 73.47\\
				%R$^3$Det$^{\dag}$ & ResNet152 & $\checkmark$ & $\checkmark$ & $\checkmark$ &  $\checkmark$ & 70.91 & 80.74 & 79.01 & 73.91\\ 
				\bottomrule
		\end{tabular}}
		\caption{Ablative study of each component in our method on the DOTA dataset. R$^3$Det$^{\dag}$ indicates that R$^3$Det with two refinement stages. BF, LK and FR denote box filtering, large kernel and feature reconstruction.}
		\label{table:ablation_study}
		\vspace{-10pt}
	\end{table}
	
	The refinement stage can be added and repeated by multiple times. The feature reconstruction process of each refinement stage is simulated as follows:
	\begin{equation}
	\begin{aligned}
	F_{i+1} = FRM(B_{i}, S_{i}, \{P_{2},...,P_{7}\})
	\label{eq:frm}
	\end{aligned}
	\end{equation}
	where $F_{i+1}$ represents the feature map of the $i+1$ stage, $B_{i}, S_{i}$ represent the bounding box and confidence score of the $ith$ stage prediction, respectively.
	
	\noindent \textbf{Discussion for comparison with RoIAlign.} The core to solve feature misalignment for FRM is feature reconstruction.
	%Deformable-based methods \cite{chen2018dual,jang2019propose,zhang2019cascade} achieve feature alignment by learning the offset between the pre-defined anchor box and the refined anchor, which which is too implicit and can not ensure that features are truely aligned. In contract, FRM clearly find the corresponding feature area by calculation. 
	Compared with RoI Align that has been adopted in many two-stage rotation detectors including R$^2$CNN and RRPN, FRM has the following differences that contribute to R$^3$Det's higher efficiency, as shown in Table \ref{table:HRSC2016_speed}.
	
	1) RoI Align has more sampling points (the default number is $7\times7\times4=196$), and reducing the sampling point greatly affects the performance of the detector. FRM only samples five feature points, about one-fortieth of RoI Align, which gives FRM a huge speed advantage.
	
	2) RoI Align need to obtain the feature corresponding to RoI \textbf{(instance level)} before classification and regression. In contrast, FRM first obtains the features corresponding to the feature points, and then reconstructs the entire feature map \textbf{(image level)}. As a result, the FRM based method can maintain a full convolution structure that leads to higher efficiency and fewer parameters, compared with the RoI Align based method that involves a fully-connected structure.
	
	\section{Experiments}
	%Tests are implemented by TensorFlow \cite{abadi2016tensorflow} on a server with GeForce RTX 2080 Ti and 11G memory.  
	%We perform experiments on both aerial benchmarks and scene text benchmarks to verify the generality of our techniques. 
	
	\subsection{Datasets and Protocls}
    The public dataset DOTA \cite{xia2018dota} is comprised of 2,806 large aerial images from different sensors and platforms. Objects in DOTA exhibit a wide variety of scales, orientations, and shapes. These images are then annotated by experts using 15 object categories. 
	The short names for categories are defined as (abbreviation-full name): PL-Plane, BD-Baseball diamond, BR-Bridge, GTF-Ground field track, SV-Small vehicle, LV-Large vehicle, SH-Ship, TC-Tennis court, BC-Basketball court, ST-Storage tank, SBF-Soccer-ball field, RA-Roundabout, HA-Harbor, SP-Swimming pool, and HC-Helicopter. 
	The fully annotated DOTA benchmark contains 188,282 instances, each of which is labeled by an arbitrary quadrilateral. 
	There are two detection tasks for DOTA: horizontal bounding boxes (HBB) and oriented bounding boxes (OBB). 
	Half of the original images are randomly selected as the training set, 1/6 as the validation set, and 1/3 as the testing set. We divide the images into $ 600 \times 600 $ subimages with an overlap of 150 pixels and scale it to $ 800 \times 800 $. With all these processes, we obtain about 27,000 patches.

	\begin{table}[tb!]
		\centering
		\resizebox{0.47\textwidth}{!}{
				\begin{tabular}{lcccccccc}
					\toprule
					\multirow{2}{*}{Method}&\multicolumn{2}{c}{FRM} & \multicolumn{3}{c}{ICDAR2015} & \multicolumn{2}{c}{HRSC2016} &UCAS-AOD \\
					\cmidrule(lr){2-3} \cmidrule(lr){4-6} \cmidrule(lr){7-8} \cmidrule(lr){9-9}
					& BF\&FR & LK & Recall & Precision & Hmean & mAP (07) & mAP (12) & mAP\\
					\midrule
					
					R$^3$Det$^*$ & & $\checkmark$ & 81.64 & 84.97 & 83.27 & 89.14 & 94.98 & 95.03 \\
					R$^3$Det & $\checkmark$ & $\checkmark$ & 83.54 & 86.43 & 84.96 (\textbf{+1.69}) & 89.26 (\textbf{+0.12}) & 96.01 (\textbf{+1.03}) & 96.17 (\textbf{+1.14}) \\

					\bottomrule
				\end{tabular}}
		\caption{Comparison between R$^3$Det$^*$ and R$^3$Det on three datasets. 07 or 12 means 2007 or 2012 evaluation metric.}
		\label{table:dataset_frm}
		\vspace{-5pt}
	\end{table}
	
	\begin{table}[tb!]
		\centering
		\resizebox{0.48\textwidth}{!}{
			\begin{tabular}{cccc}
				
				\toprule
				
				Align Mode & Feature Refinement Interpolation Formula & Feature Extraction & mAP \\
				\midrule
				
				\textbf{FRM} &\textbf{$F_{lt} * A_{rb} + F_{rt} * A_{lb} + F_{rb} * A_{lt} + F_{lb} * A_{rt}$} & \textbf{Bilinear} & \bf{66.31} \\
				 FRM & $F_{lt} * A_{lt} + F_{rt} * A_{rt} + F_{rb} * A_{rb} + F_{lb} * A_{lb}$ & Random Bilinear & 64.28 \\
			FRM & $F_{lt} * A_{lb} + F_{rt} * A_{rb} + F_{rb} * A_{rt} + F_{lb} * A_{lt}$ & Random Bilinear & 64.37 \\
			FRM & $F_{lt} * 1 + F_{rt} * 0 + F_{rb} * 0 + F_{lb} * 0$ & Quantification & 64.02 \\
				FRM & $F_{lt} * 0 + F_{rt} * 0 + F_{rb} * 1 + F_{lb} * 0$ & Quantification & 64.19 \\\hline
				deformable& - & - & 63.56\\
				\bottomrule
				
		\end{tabular}}
		\caption{Experiments with our feature alignment technique (FRM) and learnable deformable module \cite{dai2017deformable} with different interpolation formulas. Feature interpolation has location-sensitive properties.}
		\label{table:location_sensitivity}
		\vspace{-10pt}
	\end{table}
	
	The HRSC2016 dataset \cite{liu2017high} contains images from two scenarios including ships on sea and ships close inshore. All the images are collected from six famous harbors. The image sizes range from $ 300 \times 300 $ to $ 1,500 \times 900$. The training, validation and test set include 436 images, 181 images and 444 images, respectively. 
	
	ICDAR2015 \cite{karatzas2015icdar} is used in Challenge 4 of ICDAR 2015 Robust Reading Competition. It includes a total of 1,500 pictures, 1000 of which are used for training and the remaining are for testing. The text regions are annotated by 4 vertices of the quadrangle. 
	
	UCAS-AOD \cite{zhu2015orientation} contains 1,510 aerial images of approximately $ 659 \times 1,280 $ pixels, with two categories of 14,596 instances in total. In line with~\cite{azimi2018towards,xia2018dota}, we randomly select 1,110 for training and 400 for testing. 
	
	For all datasets, the models are trained by 20 epochs in total, and learning rate is reduced tenfold at 12 epochs and 16 epochs, respectively. The initial learning rates for RetinaNet is 5e-4. The number of image iterations per epoch for DOTA, ICDAR2015, HRSC2016 and UCAS-AOD are 54k, 10k, 5k and 5k, and doubled if data augmentation and
	multi-scale training are used. The experiments in this paper are initialized by ResNet50 \cite{he2016deep} by default unless otherwise specified. Weight decay and momentum are 0.0001 and 0.9, respectively. We employ MomentumOptimizer over 4 GPUs with a total of 4 images per minibatch (1 images per GPU). The anchors have areas of $32^{2}$ to $512^2$ on pyramid levels P3 to P7, respectively. At each pyramid level we use anchors at seven aspect ratios $\{1,1/2,2,1/3,3,5,1/5\}$ and three scales  $\{2^{0}, 2^{1/3}, 2^{2/3}\}$. We also add six angles $\{-90^\circ,-75^\circ,-60^\circ,-45^\circ,-30^\circ,-15^\circ\}$ for rotating anchor-based method (RetinaNet-R).
	
	\subsection{Robust Baseline Methods}
	In this paper, we use three robust baseline models with different anchor settings.
	
	\begin{table}[tb!]
		\centering
		\resizebox{0.4\textwidth}{!}{
			\begin{tabular}{cccccccc}
				\toprule
				\#Stages & Test stage & BR & SV &  LV &  SH & HA &  mAP\\
				\midrule
				
				1 & 1 & 39.25 & 63.50 & 50.68 & 65.93 & 51.93 & 62.79 \\
				2 & 2 & 42.72 & 65.81 & 72.76 & 70.14 & 56.07 & 66.31 \\ 
				3 & 3 & 45.14 & 67.09 & 73.70 & 70.21 & 56.96 & 67.29\\
				4 & 4 & 44.20 & 65.30 & 72.99 & 70.16 & 55.70 & 67.02 \\
				%5 & 5 & \\
				3 & $\overline{2-3}$ & 45.08 & 67.45 & 73.98 & 70.27 & 57.30 & 67.66 \\
				\bottomrule
				
		\end{tabular}}
		\caption{Ablation study for number of stages on DOTA dataset. $\overline{2-3}$ indicates the ensemble result, which is the collection of all outputs from refinement stages.}
		\label{table:multi-stages}
	\end{table}

	\begin{table}[tb!]
		\centering
		\resizebox{0.48\textwidth}{!}{
			\begin{tabular}{ccccc}
				
				\toprule
				Method&baseline&$-\ln(SkewIoU)$&$1-SkewIoU$& $\exp(1-SkewIoU)-1$\\
						
				\midrule
				RetinaNet-H & 62.79 & NAN & 65.06 \textbf{(+2.27)} & 65.34 \textbf{(+2.55)} \\
				R$^3$Det$^{\dag}$ & 67.66 & NAN & 68.97 \textbf{(+2.31)} & 69.50 \textbf{(+2.84)} \\
				\bottomrule
		\end{tabular}}
		\caption{Experiments with different SkewIoU functions.}
		\label{table:iou_function}
		\vspace{-10pt}
	\end{table}

\noindent \textbf{ReitnaNet-H}: The advantage of a horizontal anchor is that it can use less anchor but match more positive samples by calculating the IoU with the horizontal circumscribing rectangle of the ground truth, but it introduces a large number of non-object or regions of other objects. For an object with a large aspect ratio, its prediction rotating bounding box tends to be inaccurate, as shown in Figure \ref{fig:H_anchor_vis1}.
	
	\noindent \textbf{ReitnaNet-R}: In contrast, in Figure \ref{fig:R_anchor_vis1}, the rotating anchor avoids the introduction of noise regions by adding angle parameters and has better detection performance in dense scenes. However, the number of anchor has multiplied, about 6 times in this paper, thus making the model less efficient.
	
	\noindent \textbf{R$^3$Det$^*$}: This is a refined detector without feature refinement. Considering the number of original anchors determines the speed of the model, we adopt a progressive regression form from coarse to fine. Specifically, we first use horizontal anchor to reduce the number of anchors and increase the object recall rate in the first stage, and then use the rotating refined anchor to overcome the problems caused by dense scenes in subsequent stages, as shown in Figure \ref{fig:HR_anchor_vis1}.
	
	RetinaNet-H and RetinaNet-R have similar overall mAP (62.79\% versus 62.76\%) according to Table \ref{table:ablation_study}, while with their respective characteristics. The horizontal anchor-based approach clearly has an advantage in speed, while the rotating anchor-based method has better regression capabilities in dense object scenarios and objects with large aspect ratio, such as small vehicle, large vehicle, and ship. R$^3$Det$^*$ achieves 63.52\% performance, better than RetinaNet-H and RetinaNet-R. Although the category of dense and large aspect ratio has been improved a lot, it is still not as good as RetinaNet-R (such as LV and SH). RetianNet-R's advantages in this regard will also be reflected in Table \ref{table:HRSC2016_speed}.

	\begin{table*}[tb!]
		\centering
		\resizebox{0.98\textwidth}{!}{
			\begin{tabular}{l|lcccccccccccccccccc}
				\toprule
				& Method & Backbone & MS &  PL &  BD &  BR &  GTF &  SV &  LV &  SH &  TC &  BC &  ST &  SBF &  RA &  HA &  SP &  HC &  mAP\\
				\midrule
				\multirow{12}{*}{\rotatebox{270}{\shortstack{Two-stage}}} 
				& ICN \cite{azimi2018towards} & ResNet101 & $\checkmark$ & 81.40 & 74.30 & 47.70 & 70.30 & 64.90 & 67.80 & 70.00 & 90.80 & 79.10 & 78.20 & 53.60 & 62.90 & 67.00 & 64.20 & 50.20 & 68.20 \\
				& RADet \cite{li2020radet} & ResNeXt101 & & 79.45 & 76.99 & 48.05 & 65.83 & 65.46 & 74.40 & 68.86 & 89.70 & 78.14 & 74.97 & 49.92 & 64.63 & 66.14 & 71.58 & 62.16 & 69.09 \\
				& RoI-Transformer \cite{ding2018learning} & ResNet101 & $\checkmark$ & 88.64 & 78.52 & 43.44 & 75.92 & 68.81 & 73.68 & 83.59 & 90.74 & 77.27 & 81.46 & 58.39 & 53.54 & 62.83 & 58.93 & 47.67 & 69.56 \\
				& CAD-Net \cite{zhang2019cad} & ResNet101 & & 87.8 & 82.4 & 49.4 & 73.5 & 71.1 & 63.5 & 76.7 & \textbf{90.9} & 79.2 & 73.3 & 48.4 & 60.9 & 62.0 & 67.0 & 62.2 & 69.9 \\
				& Cascade-FF \cite{hou2020cascade} & ResNet152 & & 89.9 & 80.4 & 51.7 & 77.4 & 68.2 & 75.2 & 75.6 & 90.8 & 78.8 & 84.4 & 62.3 & 64.6 & 57.7 & 69.4 & 50.1 & 71.8	\\
				& SCRDet \cite{yang2019scrdet} & ResNet101 & $\checkmark$ & 89.98 & 80.65 & 52.09 & 68.36 & 68.36 & 60.32 & 72.41 & 90.85 & \textbf{87.94} & \textbf{86.86} & 65.02 & 66.68 & 66.25 & 68.24 & 65.21 & 72.61\\
				%& GLS-Net \cite{li2020object} & ResNet101 & & 88.65 & 77.40 & 51.20 & 71.03 & 73.30 & 72.16 & 84.68 & 90.87 & 80.43 & 85.38 & 58.33 & 62.27 & 67.58 & 70.69 & 60.42 & 72.96 \\
				& FADet \cite{li2019feature} & ResNet101 & $\checkmark$ & \textbf{90.21} & 79.58 & 45.49 & 76.41 & 73.18 & 68.27 & 79.56 & 90.83 & 83.40 & 84.68 & 53.40 & 65.42 & 74.17 & 69.69 & 64.86 & 73.28\\
				& Gliding Vertex \cite{xu2020gliding} & ResNet101 & & 89.64 & 85.00 & 52.26 & \textbf{77.34} & 73.01 & 73.14 & 86.82 & 90.74 & 79.02 & 86.81 & 59.55 & \textbf{70.91} & 72.94 & 70.86 & 57.32 & 75.02 \\
				& Mask OBB \cite{wang2019mask} & ResNeXt101 & $\checkmark$ & 89.56 & \textbf{85.95} & 54.21 & 72.90 & 76.52 & 74.16 & 85.63 & 89.85 & 83.81 & 86.48 & 54.89 & 69.64 & 73.94 & 69.06 & 63.32 & 75.33 \\
				& FFA \cite{fu2020rotation} & ResNet101 & $\checkmark$ & 90.1 & 82.7 & 54.2 & 75.2 & 71.0 & 79.9 & 83.5 & 90.7 & 83.9 & 84.6 & 61.2 & 68.0 & 70.7 & 76.0 & 63.7 & 75.7 \\
				& APE \cite{zhu2020adaptive} & ResNeXt101 & & 89.96 & 83.62 & 53.42 & 76.03 & 74.01 & 77.16 & 79.45 & 90.83 & 87.15 & 84.51 & \textbf{67.72} & 60.33 & \textbf{74.61} & 71.84 & 65.55 & 75.75 \\
				& CenterMap OBB \cite{wang2020learning} & ResNet101 & $\checkmark$ & 89.83 & 84.41 & \textbf{54.60} & 70.25 & 77.66 & 78.32 & 87.19 & 90.66 & 84.89 & 85.27 & 56.46 & 69.23 & 74.13 & 71.56 & 66.06 & 76.03\\
				\hline
				\multirow{7}{*}{\rotatebox{270}{\shortstack{Single-stage}}} 
				& IENet \cite{lin2019ienet} & ResNet101 & $\checkmark$ & 80.20 & 64.54 & 39.82 & 32.07 & 49.71 & 65.01 & 52.58 & 81.45 & 44.66 & 78.51 & 46.54 & 56.73 & 64.40 & 64.24 & 36.75 & 57.14 \\
				& PIoU \cite{chen2020piou} & DLA-34 & & 80.9 & 69.7 & 24.1 & 60.2 & 38.3 & 64.4 & 64.8 & 90.9 & 77.2 & 70.4 & 46.5 & 37.1 & 57.1 & 61.9 & 64.0 & 60.5 \\
				& P-RSDet \cite{zhou2020objects} & ResNet101 & $\checkmark$ & 89.02 & 73.65 & 47.33 & 72.03 & 70.58 & 73.71 & 72.76 & 90.82 & 80.12 & 81.32 & 59.45 & 57.87 & 60.79 & 65.21 & 52.59 & 69.82 \\
				& O$^2$-DNet \cite{wei2019oriented} & Hourglass104 & $\checkmark$ & 89.31 & 82.14 & 47.33 & 61.21 & 71.32 & 74.03 & 78.62 & 90.76 & 82.23 & 81.36 & 60.93 & 60.17 & 58.21 & 66.98 & 61.03 & 71.04 \\
				& DRN \cite{pan2020dynamic} & Hourglass104 & $\checkmark$ & 89.71 & 82.34 & 47.22 & 64.10 & 76.22 & 74.43 & 85.84 & 90.57 & 86.18 & 84.89 & 57.65 & 61.93 & 69.30 & 69.63 & 58.48 & 73.23 \\
				\cline{2-20}
				& R$^3$Det$^\dagger$ (Ours) & ResNet101 & & 88.76 & 83.09 & 50.91 & 67.27 & 76.23 & 80.39 & 86.72 & 90.78 & 84.68 & 83.24 & 61.98 & 61.35 & 66.91 & 70.63 & 53.94 & 73.79\\
				& R$^3$Det (Ours) & ResNet152 & $\checkmark$ & 89.80 & 83.77 & 48.11 & 66.77 & \textbf{78.76} & \textbf{83.27} & \textbf{87.84} & 90.82 & 85.38 & 85.51 & 65.67 & 62.68 & 67.53 & \textbf{78.56} & \textbf{72.62} & \textbf{76.47}\\
				\bottomrule
		\end{tabular}}
		\caption{Detection accuracy on different objects (AP) and overall performance (mAP) evaluation on DOTA. R$^3$Det$^{\dag}$ indicates that two refinement stages have been added. MS indicates that multi-scale training or testing is used.}
		\label{table:DOTA_OBB}
	\end{table*}
	
	\subsection{Ablation Study}
	\textbf{Feature Refinement Module.} Table \ref{table:ablation_study} shows that R$^3$Det$^*$ can improve performance by about 0.8\% which is not significant. We believe that the main reason is that the feature misalignment problem. FRM reconstructs the feature map based on the refined anchor, which increases the overall performance by 2.79\% to 66.31\% according to Table \ref{table:ablation_study}. In order to further verify the effectiveness of FRM, we have also verified it in other datasets, including the text dataset ICDAR2015, and remote sensing dataset HRSC2016 and UCAS-AOD. FRM still shows a stronger performance advantage. As shown in Table \ref{table:dataset_frm}, the FRM-based method is improved by 1.69\%, 0.12\% (1.03)\%, and 1.14\% respectively under the same experimental configuration. Compared with (deformable) learning based method, our learning-free FRM is more accurate and effective according to Table \ref{table:location_sensitivity}. When we randomly disturb the order of the four weights in the interpolation formula, the final performance of the model will be greatly reduced, rows 3-4 of Table \ref{table:location_sensitivity}. The same conclusions have also appeared in the experiments of quantitative operations, see rows 5-6 of Table \ref{table:location_sensitivity}. This phenomenon reflects the location sensitivity of the feature points and explains why the performance of the model can be greatly improved after the feature is correctly refined.
	
	\noindent \textbf{Number of Refinement Stages.} 
	Refinement strategy can significantly improve the performance of rotation detection, especially the introduction of feature refinement. Table \ref{table:ablation_study} explores the relationship between the number of refinements and model performance. R$^3$Det$^{\dag}$ has joined the two refinement stages and bring more gain. To further explore the impact of the number of stages, several experimental results are summarized in Table \ref{table:multi-stages}. Experiments show that three or more refinements will not bring additional improvements to overall performance. We also find that ensemble multi-stage results can further improve detection performance.
	
	\noindent \textbf{Approximate SkewIoU Loss.}
	We use two different detectors and three different SkewIoU functions to verify the effectiveness of the approximate SkewIoU, as shown in Table \ref{table:iou_function}. RetinaNet-based detectors will have a large number of low-SkewIoU prediction bounding box in the early stage of training, and will produce very large loss after the $\log$ function, and training is prone to non-convergence. Compared with the linear function, the derivative of the $\exp$-based function is related to SkewIoU, that is, more attention is paid to the training of difficult samples, so it has a higher performance improvement. Compare with PIoU, we can achieve considerable gains on a higher baseline and far exceed PIoU in final performance, 73.79\% versus 60.5\% as shown in Table \ref{table:DOTA_OBB}.
	
	\subsection{Comparison with the State-of-the-Art}
	%The proposed R$^3$Det is compared to state-of-the-art object detectors on three datasets: DOTA \cite{xia2018dota}, HRSC2016 \cite{liu2017high} and UCAS-AOD \cite{zhu2015orientation}.
	
	\begin{table}[tb!]
		\centering
		\resizebox{0.48\textwidth}{!}{
			\begin{tabular}{lccccc}
				\toprule
				
				Method & Backbone & Image Size & mAP (07) & mAP (12) & Speed\\
				
				\midrule
                R$^2$CNN \cite{jiang2017r2cnn} & ResNet101 & 800*800 & 73.07 & 79.73 & 5fps \\
				RC1 \& RC2 \cite{liu2017high} & VGG16 & -- & 75.7 & -- & -- \\
				RRPN \cite{ma2018arbitrary} &  ResNet101 & 800*800 & 79.08 & 85.64 & 1.5fps \\
				R$^2$PN \cite{zhang2018toward}  &  VGG16 & -- & 79.6 & -- & -- \\
				RetinaNet-H & ResNet101 & 800*800 & 82.89 & 89.27 & 14fps \\
				RRD \cite{liao2018rotation} & VGG16 & 384*384 & 84.3 & -- & -- \\
				RoI-Transformer \cite{ding2018learning} & ResNet101 & 512*800 & 86.20 & -- & 6fps \\
				Gliding Vertex \cite{xu2020gliding} & ResNet101 & -- & 88.20 & -- & -- \\
				DRN \cite{pan2020dynamic} & Hourglass104 & -- & -- & 92.70 & -- \\
				SBD \cite{liu2019omnidirectional} & ResNet50 & -- & -- & 93.70 & -- \\
				R$^3$Det$^*$ & ResNet101 & 800*800 & 89.14 & 94.98 & 4fps \\
				RetinaNet-R & ResNet101 & 800*800 & 89.18 & 95.21 & 8fps \\
				\hline
				\multirow{6}{*}{R$^3$Det} & ResNet101 & 300*300 & 87.14 & 93.22 & 18fps \\
				& ResNet101 & 600*600 & 88.97 & 94.61 & 15fps \\
				& ResNet101 & 800*800 & \textbf{89.26} & \textbf{96.01} & 12fps \\
				& MobileNetV2 & 300*300 & 77.16 & 84.31 & \bf{23fps} \\
				& MobileNetV2 & 600*600 & 86.67 & 92.83 & 20fps \\
				& MobileNetV2 & 800*800 & 88.71 & 94.45 & 16fps \\	
				\bottomrule
		\end{tabular}}
		\caption{Accuracy and speed on HRSC2016. 07 (12) means using the 2007 (2012) evaluation metric.}
		\label{table:HRSC2016_speed}
	\end{table}
	
	\noindent \textbf{Results on DOTA.} 
	We compare our results with the state-of-the-arts in DOTA as depicted in Table \ref{table:DOTA_OBB}. 
	%The results of DOTA reported here are obtained by submitting our predictions to the official DOTA evaluation server\footnote{https://captain-whu.github.io/DOTA/}. 
	Two-stage detectors are still dominant in DOTA, and the latest two-stage detection methods, such as ROI Transformer, SCRDet, and CenterMap OBB have performed well. However, they all use complex model structures in exchange for performance improvements, which are extremely low in terms of detection efficiency. The advantage of the two-stage method on the DOTA dataset lies in the multi-stage regression and the use of low-level feature maps (P2) that are friendly to small objects. Compared to all published single-stage methods, our method achieves the best performance without using multi-scale training and testing, at 73.79\%. By using a stronger backbone and multi-scale training and testing, as used in the most advanced two-stage method CenterMask OBB, R$^3$Det performs competitive performance, about 76.47\%.

	\begin{table}[tb!]
		\centering
		\resizebox{0.33\textwidth}{!}{
			\begin{tabular}{lccc}
				\toprule
				
				Method & mAP &  Plane & Car \\
				
				\midrule
				YOLOv2 \cite{redmon2017yolo9000} & 87.90 & 96.60 & 79.20 \\
				R-DFPN \cite{yang2018automatic} & 89.20 & 95.90 & 82.50 \\
				DRBox \cite{liu2017learning} & 89.95 & 94.90 & 85.00 \\ 
				S$^2$ARN \cite{bao2019single} & 94.90 & 97.60 & 92.20 \\
				RetinaNet-H & 95.47 & 97.34 & 93.60 \\
				ICN \cite{azimi2018towards} & 95.67 & - & - \\
				FADet \cite{li2019feature} & 95.71 & \textbf{98.69} & 92.72 \\
				\hline
				R$^3$Det & \textbf{96.17} & 98.20 & \textbf{94.14} \\
				\bottomrule
				
		\end{tabular}}
		\caption{Performance evaluation on UCAS-AOD datasets.}
		\label{table:UCAS-AOD}
	\end{table}
	
	\noindent \textbf{Results on HRSC2016 and UCAS-AOD.} 
	The HRSC2016 is a challenging dataset that contains lots of large aspect ratio ship instances with arbitrary orientation. We use RRPN and R$^2$CNN for comparative experiments, which are originally used for scene text detection. Experimental results show that these two methods under-perform in the remote sensing dataset, only 73.07\% and 79.08\% respectively. Although RoI Transformer achieves 86.20\% mAP, its detection speed is still not ideal, and only about 6fps without accounting for the post-processing operations. RetinNet-H, RetinaNet-R and R$^3$Det$^*$ are the three baseline models used in this paper. RetinaNet-R achieves the best detection results, around 89.14\%, which is consistent with the performance of the ship category in the DOTA dataset. This further illustrates that the rotation-based approach has advantages in large aspect ratio object detection. Under ResNet101 backbone, R$^3$Det can achieve better performance than Gliding Vertex, DRN, SDB and above methods. Besides, our method can achieve 86.67\% accuracy and 20fps speed, given MobileNetv2 \cite{sandler2018mobilenetv2} as backbone with input image size $ 600 \times 600 $. Table \ref{table:UCAS-AOD} illustrates the comparison of performance on UCAS-AOD dataset, our results are the best out of all the existing published methods, at 96.17\%.
	
	\section{Conclusion}
	We have presented an end-to-end refined single-stage detector designated for rotating objects with large aspect ratio, dense distribution and arbitrary orientations, which are common in practice like aerial, retail and scene text image. Seeing the shortcoming of feature misalignment in the current refined single-stage detector, we design a feature refinement module to improve detection performance. The key idea of FRM is to re-encode the position information of the current refined bounding box to the corresponding feature points through pixel-wise feature interpolation to achieve feature reconstruction and alignment. For more accurate rotation estimation, an approximate SkewIoU loss is proposed to solve the problem that the calculation of SkewIoU is not derivable. We perform careful ablation studies and comparative experiments on multiple rotation detection datasets including DOTA, HRSC2016, UCAS-AOD, and ICDAR2015, and demonstrate that our method achieves state-of-the-art detection accuracy with high efficiency.
	
\section{Supplementary Material}
	
	\begin{figure}[!tb]
		\begin{center}
			\includegraphics[width=0.95\linewidth]{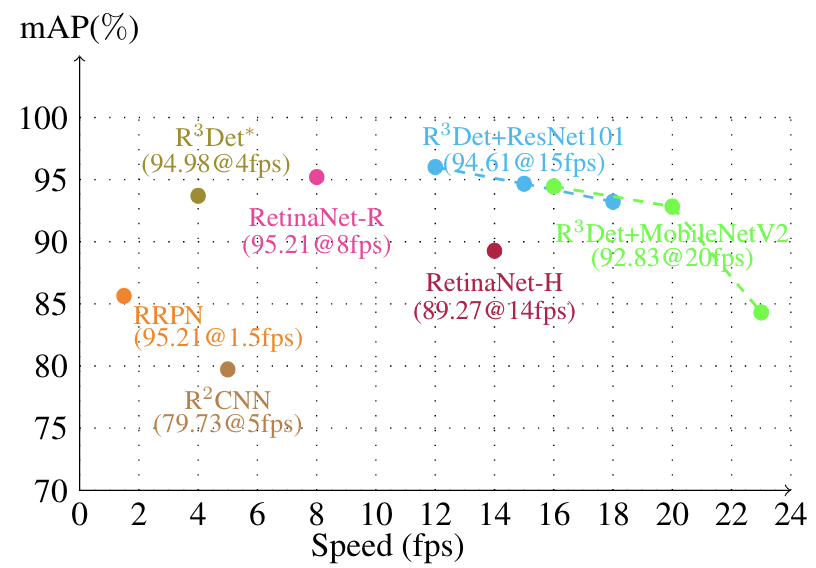}
		\end{center}
		\vspace{-10pt}
		\caption{Performance versus speed on HRSC2016 \cite{liu2017high} dataset. As can be seen, our algorithm significantly surpasses competitors in accuracy, whilst running very fast.}
		\label{fig:hrsc2016_speed}
		\vspace{-8pt}
	\end{figure}

\subsection{Speed Comparison}
    Due to the high-resolution test images and extra processing such as image cropping and test results merging, DOTA is not suitable for speed comparison. Therefore, we compare the speed and accuracy with the other six methods on the HRSC2016 dataset under the same test environment. The time of post process (i.e. R-NMS) is included. We also explore the impact of different backbones and image sizes on the per performance of the proposed model. The detailed experimental results are shown in Figure \ref{fig:hrsc2016_speed}. Our method can achieve 92.83\% accuracy and 20fps speed, given MobileNetv2 as backbone with input image size $ 600 \times 600 $. The HRSC2016 dataset contains only one category, so the speed comparsion is not obvious between RetinaNet-R and R$^3$Det according to $(5+C) \cdot A$, but the gap will gradually widen as the number of categories increases.
    
\subsection{Results on ICDAR2015.} 
	Scene text detection is also one of the main application scenarios for rotation detection. As shown in Table \ref{table:icdar2015}, our method achieves 84.96\% while maintaining 13.5fps in the ICDAR2015 dataset, better than most mainstream algorithms except for FOTS, which adds a lot of extra training data (like ICDAR 2017 MLT \cite{nayef2017icdar2017}) and uses large test image. With these heavy settings, our method R$^3$Det$^\ddagger$ achieved 89.21\% and still maintains 9fps. The experimental results still show that the proposed techniques are general that can be useful for both aerial images and scene text images.

\begin{table}[tb!]
		\centering
		\resizebox{0.48\textwidth}{!}{
				\begin{tabular}{lccccccc}
					\hline
					\multirow{2}{*}{Method}&\multicolumn{2}{c}{FRM}& \multirow{2}{*}{Recall} & \multirow{2}{*}{Precision} & \multirow{2}{*}{Hmean} & \multirow{2}{*}{Res.} & \multirow{2}{*}{FPS} \\
					\cline{2-3}
					& BF\&FR & LK & & & & & \\
					\hline
					CTPN \cite{tian2016detecting}& & & 51.56 & 74.22 & 60.85 & -- & --  \\
					SegLink \cite{shi2017detecting} & & & 76.80 & 73.10 & 75.00 & -- & --  \\
					RRPN \cite{ma2018arbitrary} & & & 82.17 & 73.23 & 77.44 & -- & $<$1 \\
					EAST \cite{Zhou2017EAST} & & & 78.33 & 83.27 & 80.72 & 720p & 13.2 \\
					DDR \cite{he2017deep} & & & 80.00 & 82.00 & 81.00 & -- & $<$1 \\
					R$^2$CNN \cite{jiang2017r2cnn} & & & 79.68 & 85.62 & 82.54 & 720p & 0.44 \\
					FOTS RT \cite{liu2018fots} & & & 85.95 & 79.83 & 82.78 & 720p & \bf{24} \\
					FOTS \cite{liu2018fots} & & & \bf{91.0} & 85.17 & 87.99 & 1260p & 7.8 \\
					\hline
					R$^3$Det$^*$ & $\checkmark$ & & 81.64 & 84.97 & 83.27 & 720p & 4 \\
					R$^3$Det & $\checkmark$ & $\checkmark$ & 83.54 & 86.43 & 84.96 & 720p & 13.5 \\	
					R$^3$Det$^\ddagger$ & $\checkmark$ & $\checkmark$ & 89.45 & \textbf{88.97} & \textbf{89.21} & 1024p & 9 \\		
					\hline
				\end{tabular}}
		\caption{Performance on ICDAR2015. Tick means the module is enabled}
		\label{table:icdar2015}
	\end{table}
	
	\subsection{FRM is Suitable for Rotation Detection}
\begin{figure}[!tb]
    		\centering
    		\subfigure[Horizontal detection]{
			\begin{minipage}[t]{0.46\linewidth}
				\centering
				\includegraphics[width=0.98\linewidth]{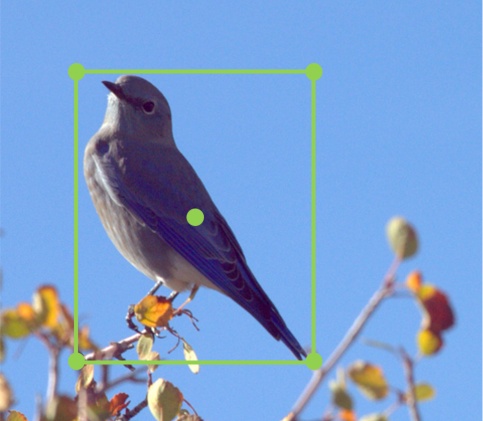}
			\end{minipage}
			\label{fig:5points1}
		}
		\subfigure[Rotation detection]{
			\begin{minipage}[t]{0.46\linewidth}
				\centering
				\includegraphics[width=0.98\linewidth]{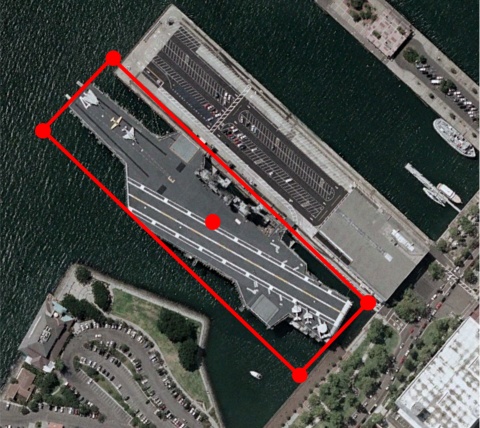}
			\end{minipage}%
			\label{fig:5points2}
		}
    	\caption{Schematic diagram of sampling points for FRM in horizontal detection and rotation detection. The sample points for horizontal detection are significantly further away from the object, and the sampling points for rotation detection are tighter.}
    	\label{fig:5points}
    \end{figure}

    	When applying FRM to horizontal detection, the feature vectors of the four corner points (green points in Figure \ref{fig:5points1}) obtained in FRM are likely to be far from the object,  resulting in very inaccurate features being sampled. However, in the rotation detection task, the four corner points (red points in Figure \ref{fig:5points2}) of the rotating bounding box are very close to the object. We have experimented on COCO dataset and the results are not satisfactory, but we have achieved considerable gains on many rotating datasets.
    
    \subsection{Training Loss Curve}
\begin{figure}[!tb]
		\centering
		\centering
		\includegraphics[width=.46\textwidth]{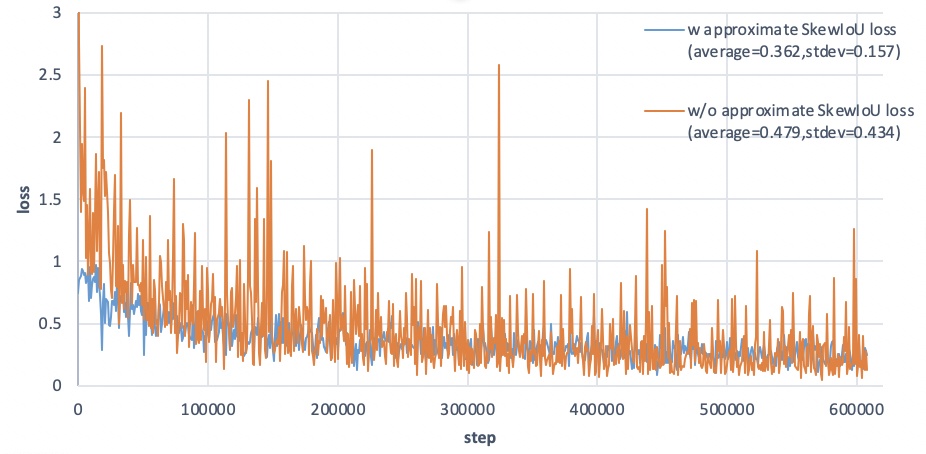}
		\caption{Training loss curve after using approximate SkewIoU loss and Smooth L1 loss.}
		\label{fig:skewiou_loss}
    \end{figure}

        Figure \ref{fig:skewiou_loss} shows the training loss curve after using approximate SkewIoU loss and Smooth L1 loss. It can be clearly seen from the variance and mean of the two curves that the training is more stable after using approximate SkewIoU loss.
	
    \subsection{Visualization on Different Datasets}
\begin{figure}[!tb]
		\centering
		\subfigure[]{
			\begin{minipage}[t]{0.46\linewidth}
				\centering
				\includegraphics[width=0.98\linewidth]{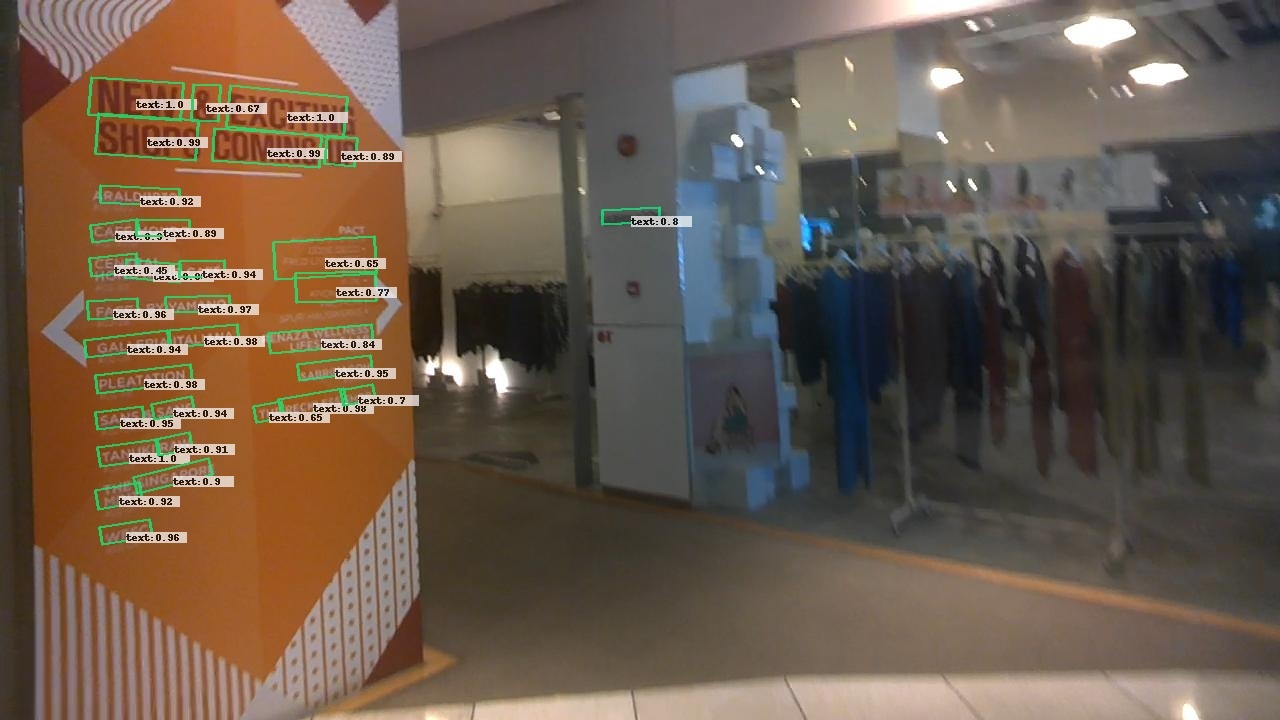}
			\end{minipage}%
			\label{fig:icdar2015_486}
		}
		\subfigure[]{
			\begin{minipage}[t]{0.46\linewidth}
				\centering
				\includegraphics[width=0.98\linewidth]{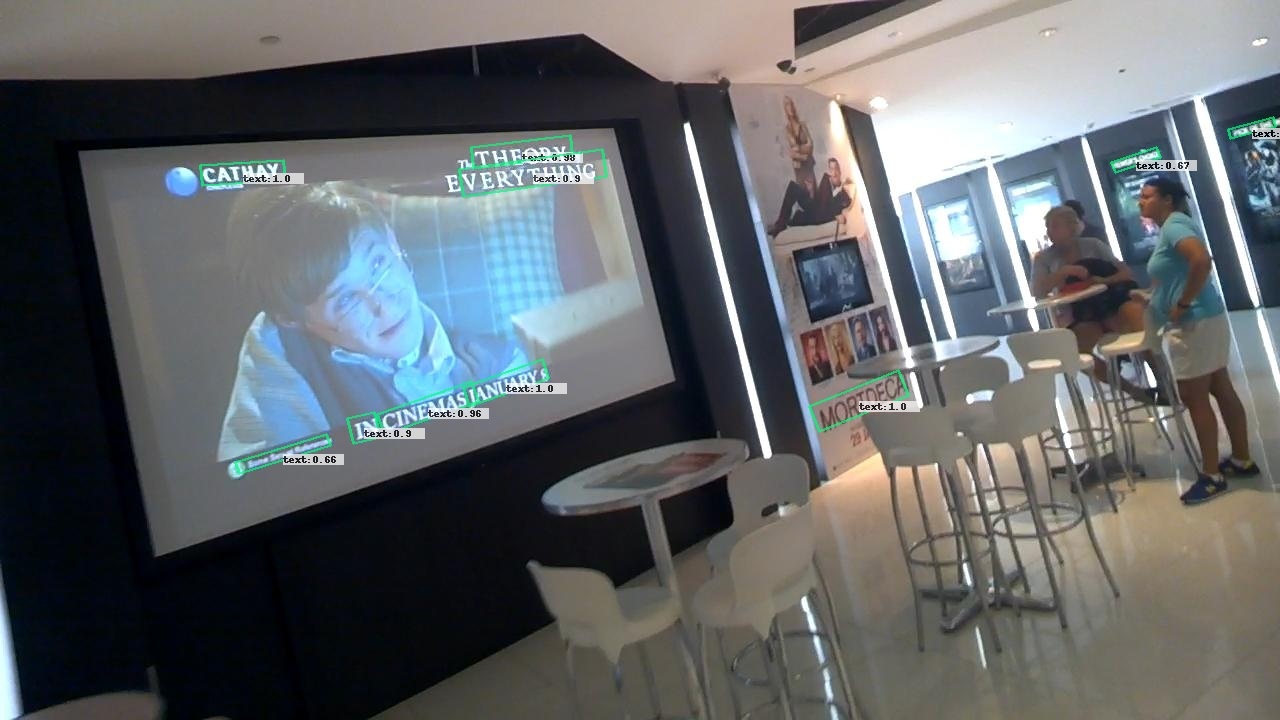}
			\end{minipage}%
			\label{fig:icdar2015_453}
		}\\
		\subfigure[]{
			\begin{minipage}[t]{0.46\linewidth}
				\centering
				\includegraphics[width=0.98\linewidth]{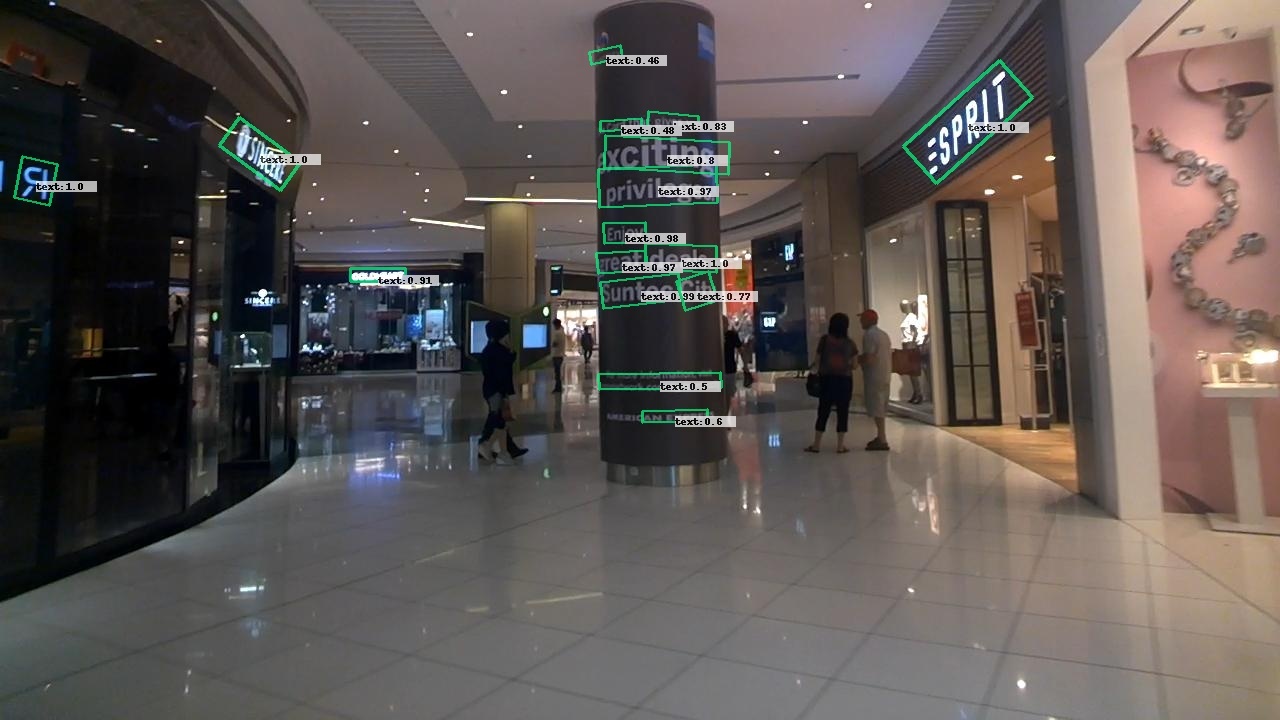}
			\end{minipage}%
			\label{fig:icdar2015_121}
		}
		\subfigure[]{
			\begin{minipage}[t]{0.46\linewidth}
				\centering
				\includegraphics[width=0.98\linewidth]{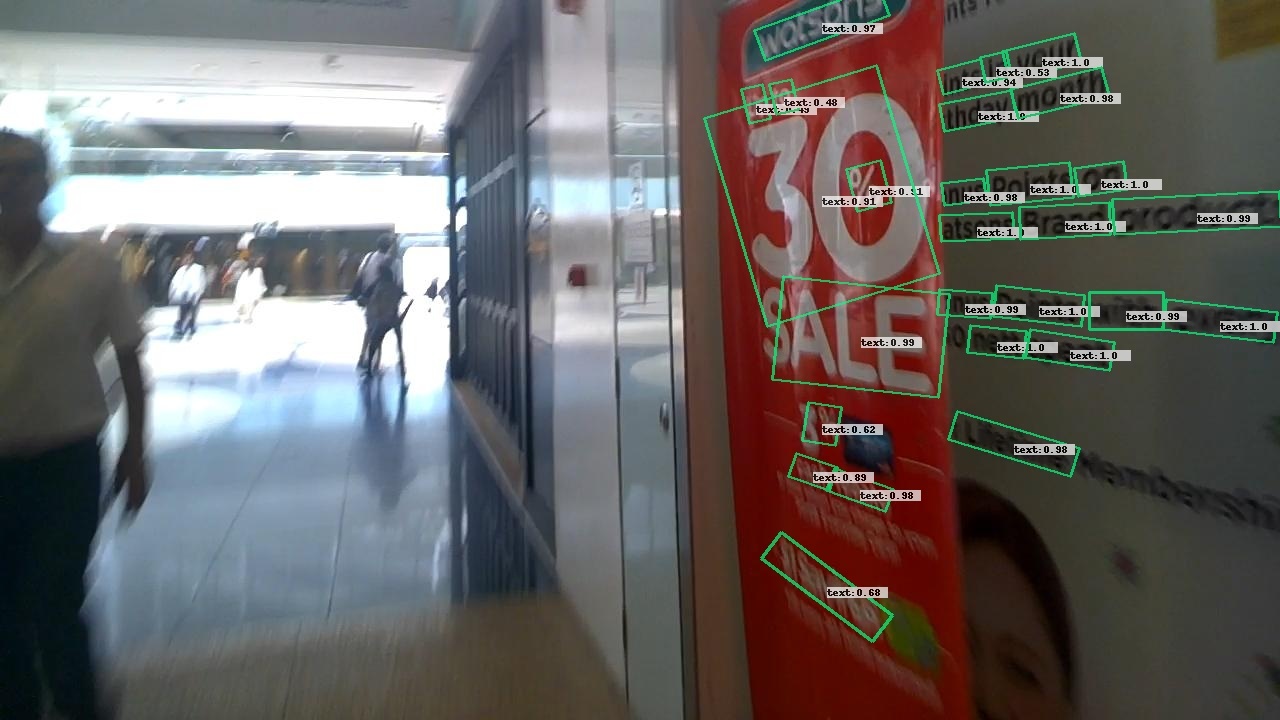}
			\end{minipage}%
			\label{fig:icdar2015_82}
		}
        \caption{Text detection results on the ICDAR2015 benchmarks.}
		\label{fig:icdar2015}
		\vspace{-8pt}
	\end{figure}

\begin{figure}[!tb]
		\centering
		\subfigure[]{
			\begin{minipage}[t]{0.46\linewidth}
				\centering
				\includegraphics[width=0.98\linewidth]{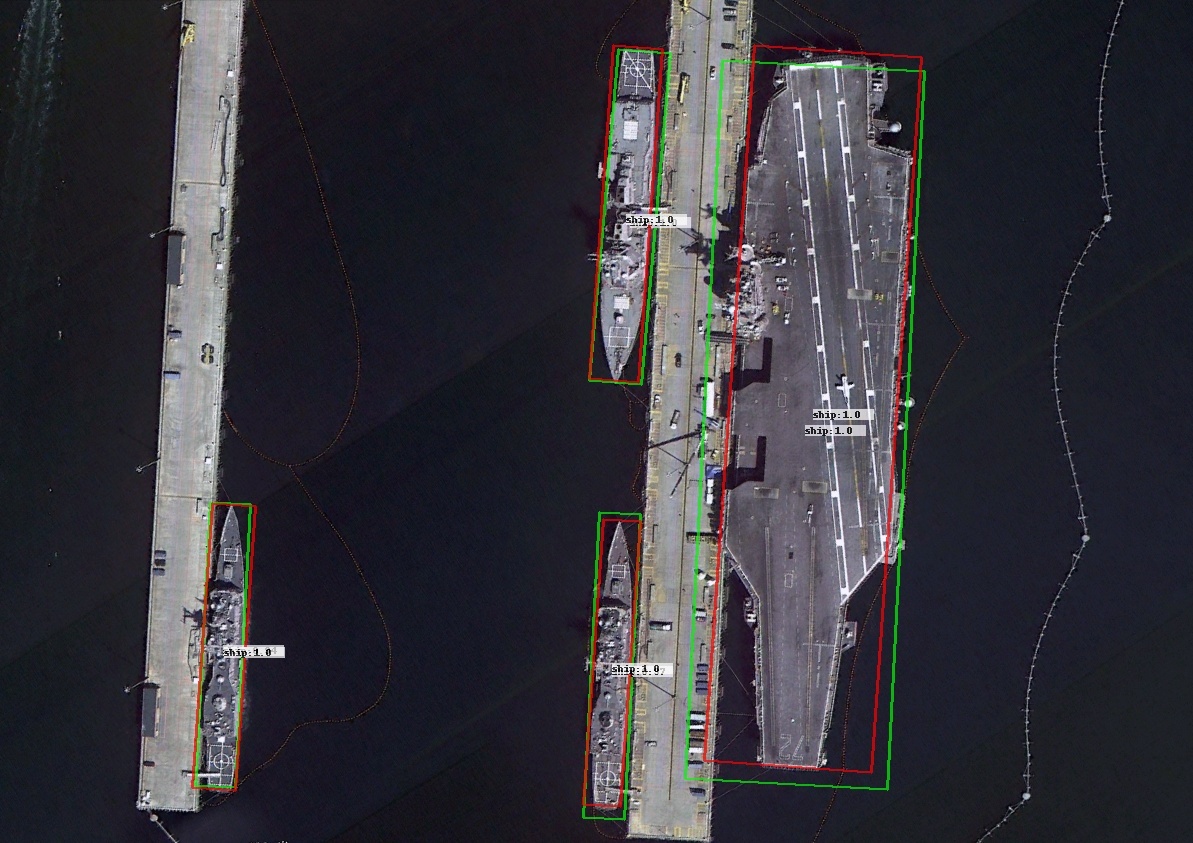}
			\end{minipage}%
			\label{fig:100000648}
		}
		\subfigure[]{
			\begin{minipage}[t]{0.46\linewidth}
				\centering
				\includegraphics[width=0.98\linewidth]{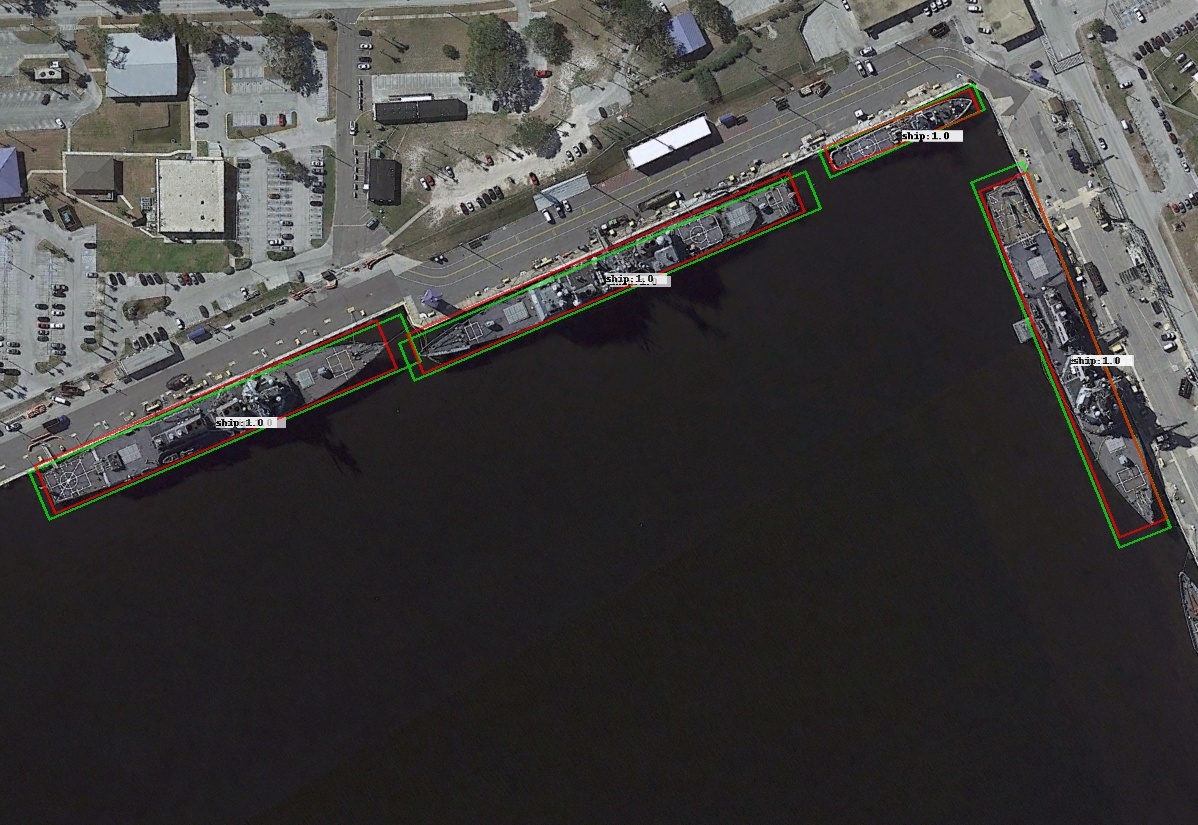}
			\end{minipage}%
			\label{fig:100000728}
		}\\
		\subfigure[]{
			\begin{minipage}[t]{0.46\linewidth}
				\centering
				\includegraphics[width=0.98\linewidth]{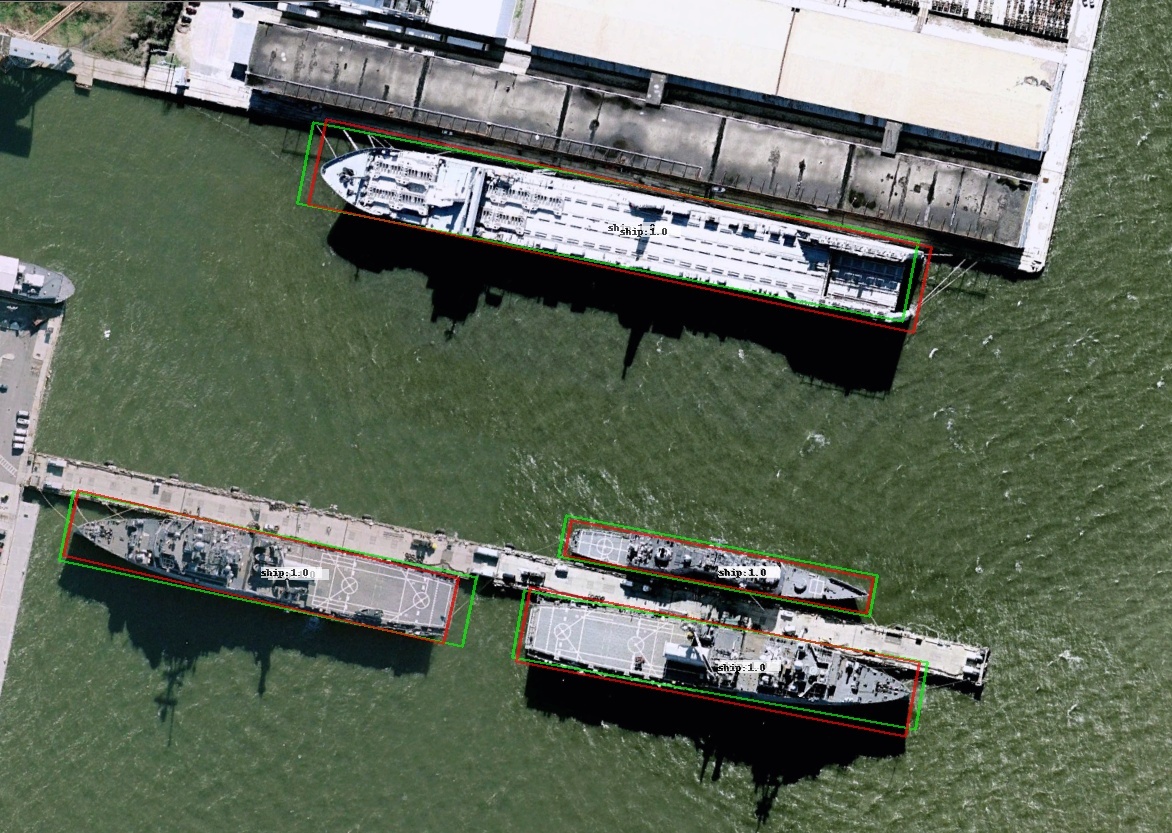}
			\end{minipage}%
			\label{fig:100001075}
		}
		\subfigure[]{
			\begin{minipage}[t]{0.46\linewidth}
				\centering
				\includegraphics[width=0.98\linewidth]{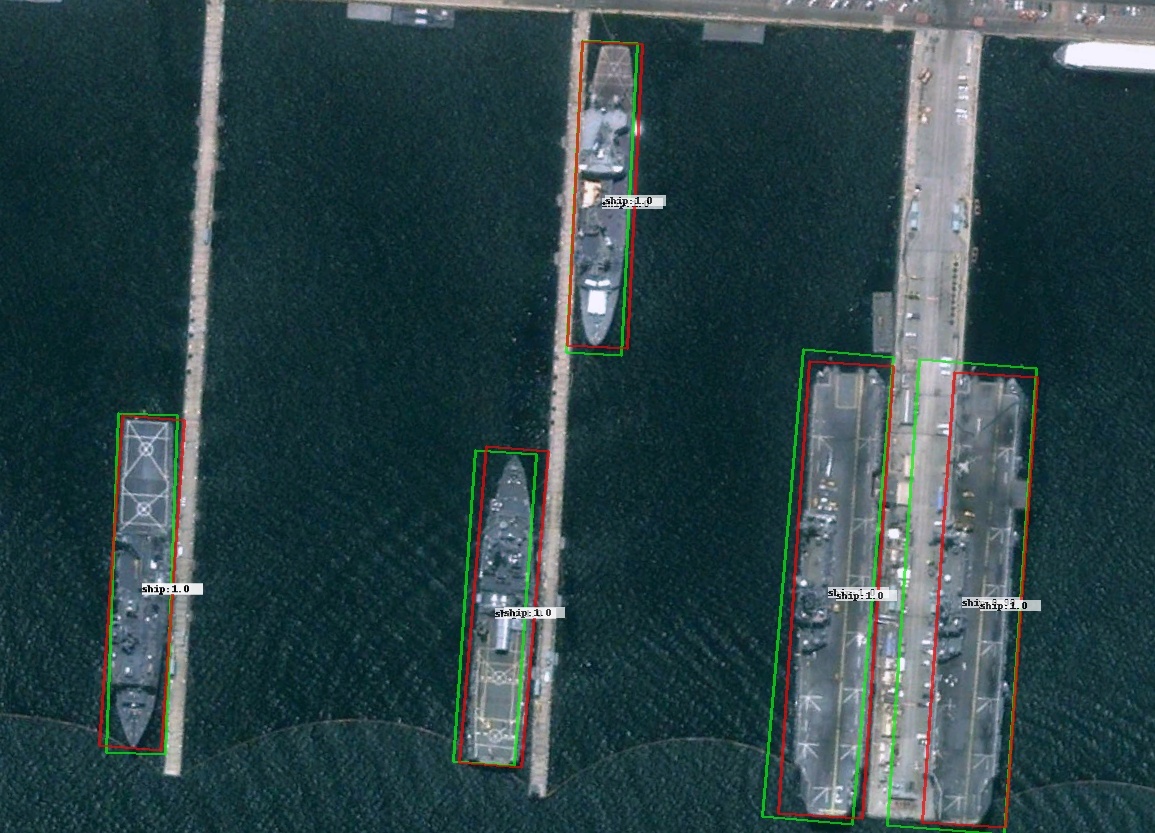}
			\end{minipage}%
			\label{fig:100001635}
		}
		\caption{Ship detection results on the HRSC2016 benchmarks. The red and green bounding box indicate the ground truth and prediction box, respectively.}
		\label{fig:hrsc2016}
	\end{figure}

\begin{figure*}[!tb]
    	\centering
    	
    	\subfigure[BC and TC]{
			\begin{minipage}[t]{0.26\linewidth}
				\centering
				\includegraphics[width=.98\linewidth, height=5.8cm]{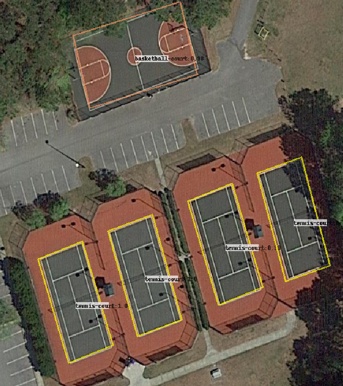}
			\end{minipage}%
			\label{fig:BC_TC}
		}
		\subfigure[SBF, GTF, TC and SP]{
			\begin{minipage}[t]{0.22\linewidth}
				\centering
				\includegraphics[width=0.98\linewidth, height=5.8cm]{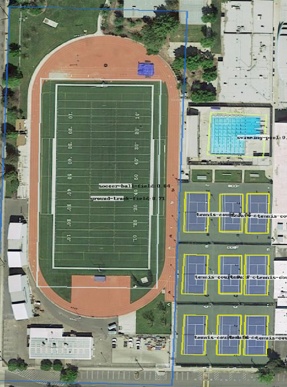}
			\end{minipage}%
			\label{fig:SBF_GTF_TC_SP}
		}
		\subfigure[HA]{
			\begin{minipage}[t]{0.22\linewidth}
				\centering
				\includegraphics[width=0.98\linewidth, height=5.8cm]{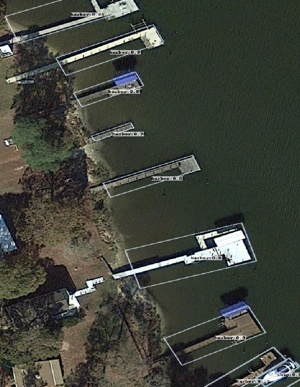}
			\end{minipage}%
			\label{fig:HA}
		}
		\subfigure[HA and SH]{
			\begin{minipage}[t]{0.22\linewidth}
				\centering
				\includegraphics[width=0.98\linewidth, height=5.8cm]{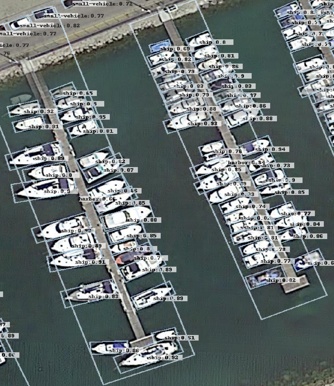}
			\end{minipage}%
			\label{fig:HA_SH}
		}\\
		\subfigure[SP]{
			\begin{minipage}[t]{0.23\linewidth}
				\centering
				\includegraphics[width=0.98\linewidth, height=4.2cm]{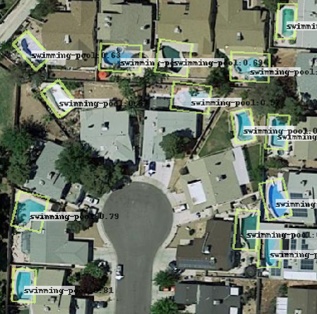}
			\end{minipage}%
			\label{fig:SP}
		}
		\subfigure[RA and SV]{
			\begin{minipage}[t]{0.23\linewidth}
				\centering
				\includegraphics[width=0.98\linewidth, height=4.2cm]{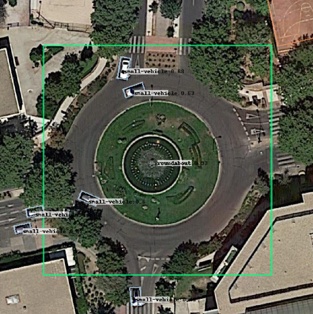}
			\end{minipage}%
			\label{fig:RA_SV}
		}
		\subfigure[ST]{
			\begin{minipage}[t]{0.23\linewidth}
				\centering
				\includegraphics[width=0.98\linewidth, height=4.2cm]{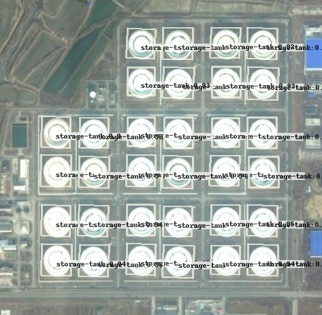}
			\end{minipage}%
			\label{fig:ST}
		}
		\subfigure[BD and RA]{
			\begin{minipage}[t]{0.23\linewidth}
				\centering
				\includegraphics[width=0.98\linewidth, height=4.2cm]{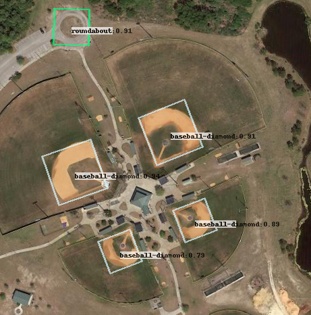}
			\end{minipage}%
			\label{fig:BD_RA}
		}\\
		\subfigure[SV and LV]{
			\begin{minipage}[t]{0.28\linewidth}
				\centering
				\includegraphics[width=0.98\linewidth, height=5.0cm]{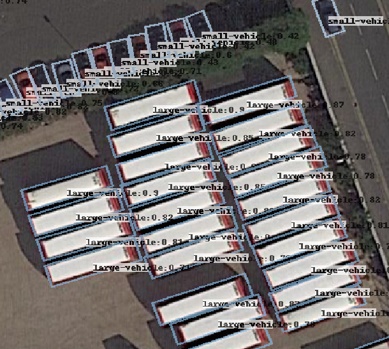}
			\end{minipage}%
			\label{fig:SV_LV}
		}
		\subfigure[PL and HC]{
			\begin{minipage}[t]{0.28\linewidth}
				\centering
				\includegraphics[width=0.98\linewidth, height=5.0cm]{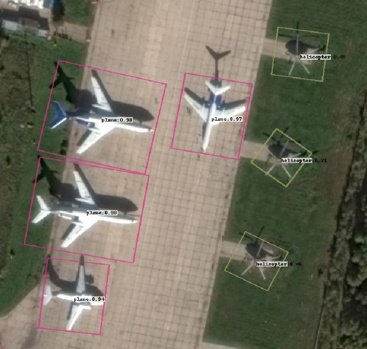}
			\end{minipage}%
			\label{fig:PL_HC}
		}
		\subfigure[BR]{
			\begin{minipage}[t]{0.38\linewidth}
				\centering
				\includegraphics[width=0.98\linewidth, height=5.0cm]{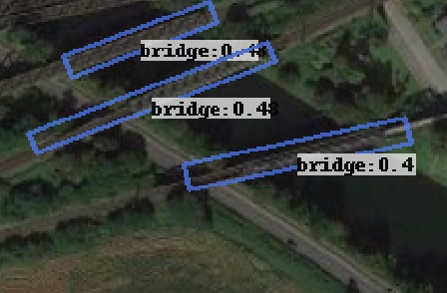}
			\end{minipage}%
			\label{fig:BR}
		} 
    	\caption{Detection results on the OBB task on DOTA. Our method performs better on those with large aspect ratio , in arbitrary direction, and high density.}
    	\label{fig:DOTA}
    \end{figure*}

    	We visualize the detection results of R$^3$Det on different types of datasets, including remote sensing datasets (Figure \ref{fig:DOTA} and Figure \ref{fig:hrsc2016}), and scene text datasets (Figure \ref{fig:icdar2015}).
    	
	\section*{Acknowledgment}
	This research was partially supported by China Major State Research Development Program (2018AAA0100704), NSFC (61972250, U19B2035). The author Xue Yang is supported by Wu Wen Jun Honorary Doctoral Scholarship, AI Institute, Shanghai Jiao Tong University.
	
	\section{Broad Societal Implications}
	This paper aims to advance the technology of rotation detection, which has wide applications in face, remote sensing, scene text and retail scene. The more accurate and efficient detection can help people more conveniently and cost-effectively record the key information while in the mean time, the privacy of individuals may be put at risk. Hence we shell take additional measures to protect privacy along the development of such technology.
	
{\small
\bibliographystyle{ieee_fullname}
\bibliography{egbib}
}

\end{document}